\documentclass[a4paper,11pt]{article}

\usepackage[utf8]{inputenc}
\usepackage[T1]{fontenc}
\usepackage{fouriernc}
\usepackage{amsthm,amssymb,amsfonts,amsmath,color,graphicx,rotating,float,url,bbm,multirow,caption,subfig,placeins,xspace,eucal}

\usepackage[boxruled]{algorithm2e}
\SetAlFnt{\scriptsize}
\SetAlCapFnt{\scriptsize}
\SetAlCapNameFnt{\scriptsize}

\makeatletter
\renewcommand{\algocf@caption@boxruled}{%
  \hrule
  \hbox to \hsize{%
    \vrule\hskip-0.4pt
    \vbox{   
       \vskip\interspacetitleboxruled%
       \unhbox\algocf@capbox\hfill
       \vskip\interspacetitleboxruled
       }%
     \hskip-0.4pt\vrule%
   }\nointerlineskip%
}%
\makeatother

\usepackage[numbers,square,sort]{natbib}
\usepackage[english]{babel}

\usepackage[pagebackref = true]{hyperref}
\hypersetup{
  colorlinks = true,
  urlcolor = blue, 
  linkcolor = blue,
  citecolor = red,
  pdftitle = {PAC-Bayesian High Dimensional Bipartite Ranking - \today},
  pdfauthor = {Benjamin Guedj and Sylvain Robbiano},
  pdfsubject = {}
}

\def\C{\mathcal{C}}
\def\V{\mathbb{V}}
\def\er{{\cal E}}
\def\ern{{\cal E}_n}

\def\R{\mathbb{R}}
\def\bx{\mathbf{x}}
\def\mp{\mathcal{M}_{\pi}}
\def\bX{\mathbf{X}}
\def\1{\mathbbm{1}}
\def\m{\mathbf{m}}
\def\PP{\mathbb{P}}
\def\e{\varepsilon}
\def\proj{\mathrm{proj}}
\def\ie{\emph{i.e.}}
\def\MP{\mathcal{M}_{+,\pi}^1}

\newcommand{\E}{\mathbb{E}}
\newcommand{\KL}{\mathcal{K}}
\newcommand{\roc}{{\rm  ROC}}
\newtheorem{theo}{Theorem}
\newtheorem{lemma}{Lemma}
\newtheorem{coro}{Corollary}
\newtheorem{cond}{Assumption}

\addto\extrasenglish{%
    
}
\usepackage{pifont}
\newcommand{\cmark}{\ding{51}}
\makeatletter
\newcommand{\vast}{\bBigg@{4}}
\newcommand{\Vast}{\bBigg@{5}}
\makeatother

\title{PAC-Bayesian High Dimensional Bipartite Ranking}
\author{Benjamin Guedj\footnote{Modal project-team, Inria, France.\ \href{mailto:benjamin.guedj@inria.fr}{benjamin.guedj@inria.fr}} \ and Sylvain Robbiano\footnote{Department of Statistical Science, UCL, United Kingdom.\ \href{mailto:sylvain.robbiano@gmail.com}{sylvain.robbiano@gmail.com}}}
\date{\today}

\begin{document}
\maketitle

\vspace{-.5cm}
\begin{abstract}
\noindent  This paper is devoted to the bipartite ranking problem, a classical statistical learning task, in a high dimensional setting. We propose a scoring and ranking strategy based on the PAC-Bayesian approach. We consider nonlinear additive scoring functions, and we derive non-asymptotic risk bounds under a sparsity assumption. In particular, oracle inequalities in probability holding under a margin condition assess the performance of our procedure, and prove its minimax optimality. An MCMC-flavored algorithm is proposed to implement our method, along with its behavior on synthetic and real-life datasets.
\smallskip


\end{abstract}



\section{Introduction}

The bipartite ranking  problem appears in various application
domains such as medical diagnostic, information retrieval, signal
detection. This supervised learning task consists in building a
so-called scoring function that order the (high dimensional) observations in the
same fashion as the (unknown) associated labels. In that sense, the global problem of bipartite ranking (ordering a set observations) includes the local classification task (assigning a label to each observation). Indeed, once a proper scoring function is defined, classification amounts to choosing a threshold, assigning data points to either class depending on whether their score is above or below that threshold.
\medskip

The quality of a
scoring function is usually assessed through its $\roc$ curve \citep[Receiver Operating Characteristic, see][]{GreSwe66}
and it is shown that maximizing this visual tool is
equivalent to solving the bipartite ranking problem
\citep[see for example, Proposition 6 in][]{CV09ieee}. Due to the functional nature of the $\roc$ curve, it is useful to substitute a proxy: maximizing the Area Under
the $\roc$ Curve (AUC), or equivalently minimizing the pairwise
risk. Following that idea, several classical algorithms in classification
have been extended to the case of bipartite ranking, such as Rankboost
\citep[][]{FISS03} or RankSVM \citep[][]{Rakot1}. Several authors
have considered theoretical aspects of this problem. In
\cite{AGHHPR05}, the authors investigate the difference between the
empirical AUC and the true AUC and produce a concentration
inequality assessing that as the number of observations increases,
the empirical AUC tends to the true AUC, allowing for empirical risk minimization (ERM) approaches to tackle the bipartite ranking problem. This
strategy has been explored by \cite{CLV08}. Assuming that the true scoring function is in a Vapnik-Cervonenkis class of finite dimension,
combined with a low noise condition, the authors prove that the minimizer of the empirical
pairwise risk achieves fast rates of convergence. More recently, the
bipartite ranking problem has been tackled from a nonparametric angle
and \cite{CRICML} proved that a plug-in estimator of the regression function can attain minimax
fast rates of convergence over H\"older class. In
order to obtain an adaptive estimator to the low noise and
to the H\"older parameters, an aggregation procedure based on
exponential weights has been proposed by \cite{REJS} and the author
shows adaptive fast rate upper bounds. However, the rates of
convergence depend on the dimension of the features space and in many
applications the optimal scoring function depends on a small number of
the features, suggesting a sparsity assumption. As a matter of fact, the problem of sparse bipartite ranking has been studied
by \cite{Li13}. The authors obtain general PAC results for the Gibbs estimator and deduce an oracle bound for linear scoring function under mild assumptions. In this paper, we consider the case of non-parametric scoring functions that can be sparsely decomposed in an additive way with respect to the covariates. Our results hold with hypothesis tailored for the ranking case and a specific set of candidate scoring function that allow for non-parametric rates.
\medskip

To do so, we design an aggregation strategy which heavily relies on the PAC-Bayesian paradigm (the acronym PAC stands for \emph{Probably Approximately Correct}). In our setting, the PAC-Bayesian approach delivers a random estimator (or its expectation) sampled from a (pseudo) posterior distribution which exponentially penalizes the AUC risk.
The 
PAC-Bayesian theory originates in the two seminal papers
\cite{SW1997} and \cite{McA1999}. The first PAC-Bayesian bounds consisted in data-dependent empirical inequalities for Bayesian-flavored estimators.
This strategy has been extensively formalized in the
context of classification by \cite{Cat2004,Cat2007} and
regression by \cite{Aud2004,Alq2008}.
PAC-Bayesian techniques have proven useful to study the convergence rates of Bayesian learning methods.
More recently, these methods have been studied under the scope of high dimensionality (typically with a sparsity assumption): see for example \cite{DT2008,DT2012,AL2011,AB2013,GA2013}. The main message of
these works is that PAC-Bayesian aggregation with
a properly chosen prior is able to deal effectively with the sparsity
issue in a regression setting under the $\ell^{2}$ loss. The purpose of
the present paper is to extend the use of such techniques to the case of bipartite ranking. Note that in a work parallel to ours, \cite{RACL2014} also use a PAC-Bayesian machinery for bipartite ranking. Their framework is close to the one developed in this paper, however the authors focus on linear scoring function and produce oracle inequality in expectation. The work presented in the present paper is more general as we consider nonlinear scoring functions and derive non-asymptotic risk bounds in probability.
\medskip

Our procedure relies on the construction of a high dimensional yet sparse (pseudo-)posterior distribution (Gibbs distribution, introduced in \autoref{S:notation}). Most of the aforecited papers studying PAC-Bayesian strategies rely on Monte Carlo Markov Chain (MCMC) algorithms to sample from this target distribution. However, very few discuss the practical implementation of an MCMC algorithm in the case of (possibly very) high dimensional data, to the notable exception of \cite{AB2013}, \cite{GA2013} (both through MCMC) and \cite{RACL2014} (Sequential Monte Carlo).
\medskip

We adapt the
point of view presented in \cite{GA2013} and implemented in the R
package \texttt{pacbpred} \citep{Gue2013a}, which is inspired
by \cite{CC1995,PD2012}. The key idea is to define a
neighborhood relationship between visited models, promoting local
moves of the Markov chain. This approach has the merit of being easily implementable, and adapts well to our will to promote sparse scoring functions. Indeed, the Markov chain will mostly visit low dimensional models, ensuring sparse scoring predictors as outputs. As emphasized in the following, this choice leads to nice performance when a sparse additive representation is a good approximation of the optimal scoring function.
\medskip

The paper is structured as follows. We
introduce the notation and setting in \autoref{S:notation} and we present our PAC-Bayesian estimation strategy for the bipartite ranking
problem in \autoref{sec:procedure}. \autoref{S:theory} contains the core of our contribution. We gather here all our theoretical results in the form of oracle inequalities in probability. Risk bounds are presented to assess the merits of our procedure and exhibit explicit non-parametric rates of convergence. To illustrate the practical potential of PAC-Bayesian high dimensional binary ranking, \autoref{S:mcmc} presents our MCMC algorithm to compute our estimator, coupled with numerical results on both synthetic and real-life datasets. Conclusive comments on both theoretical and practical merits of our work are summed up in \autoref{S:conclusion}. Finally, proofs of the original results claimed in the paper are gathered in \autoref{S:proof} for the sake of clarity.

\section{Notation}\label{S:notation}

Adopting the notation $\bX=(X_1,\dots,X_d)$, we let $(\bX,Y)$ be a random variable taking its values in
$\R^d\times\{\pm 1\}$. We let $\PP=(\mu,\eta)$ denote the distribution of $(\bX,Y)$, where $\mu$ is the marginal distribution of $\bX$, and $\eta(\cdot)=\PP[Y=1|\bX=\cdot]$. Our goal is to solve the bipartite ranking problem, \ie, ordering the features space $\R^d$ to preserve the orders induced by the labels. In other words, our goal is to design an order relationship on $\R^d$ which is consistent with the order on $\{\pm 1\}$: when given a new pair of points $(\bX,Y)$ and $(\bX^\prime,Y^\prime)$ drawn from $\PP$, order $\bX$ then $\bX^\prime$ iff $Y<Y^\prime$.
\medskip

A natural way to build up such an order relation on $\R^d$ is to transport the usual order on the real line onto $\R^d$ through a (measurable) \emph{scoring function}  $s\colon \R^d\rightarrow \R$ such that for any $(\bx,\bx')\in \R^d\times\R^d$, we have $\bx\preceq_s \bx'\Leftrightarrow s(\bx)\leq s(\bx').$

It is tempting to try and mimic the sorting performance of the unknown regression function $\eta$, which is clearly optimal.
The bipartite ranking problem may now be rephrased as building a scoring function $s$  such that, for any pair $(\bx,\bx^\prime)\in \R^d\times\R^d$, $s(\bx)\leq
s(\bx^\prime)\Leftrightarrow \eta(\bx)\leq \eta(\bx^\prime)$.
From a statistical perspective, our goal is to learn such a scoring function using a $n$-sample $\mathcal{D}_n=\{(\bX_i,Y_i)\}_{i=1}^n$ consisting in i.i.d. replications of $(\bX,Y)$.
\medskip

To assess the theoretical quality of a scoring function, it is natural to consider the ranking risk, based on the pairwise classification loss, defined as follows: let $(\bX,Y)$ and $(\bX^\prime,Y^\prime)$ be two independent variables drawn from some distribution $\PP$. The ranking risk of some scoring function $s$ is 
\begin{equation*}
L(s)=\mathbb{P}\left[(s(\bX)-s(\bX^\prime)) (Y^\prime-Y)<0\right].
\end{equation*}
This quantity is closely related to the AUC, \begin{equation*}
\textrm{AUC}(s)=\mathbb{P}[s(\bX)<s(\bX^\prime)\vert Y=-1,Y^\prime=+1],
\end{equation*}
as pointed out by \cite{CLV08}. Note that the authors have proved that the optimal scoring function for the ranking risk is the posterior distribution $\eta$, which is obviously unknown to the statistician. The problem of AUC estimation consists in looking at how far an estimator of the AUC is of the true AUC. In this paper we use the empirical ranking risk which is closely related to the non-parametric estimator of the AUC. The proximity of this estimator to the true ranking risk is controlled through the exponential inequality in \autoref{lemmacond} for the general case and \autoref{coro2} under the margin condition.
\medskip

The approach adopted in this paper consists in providing bounds on the \emph{excess ranking risk}, defined as $$\mathcal{E}(\cdot) = L(\cdot)-L^\star,$$ where $L^\star:=L(\eta)$. Since $\eta$ is the minimizer of $L$, $\er$ is a positive valued function.
In \cite{CLV08}, it is shown that the excess ranking risk may be reformulated as
\begin{equation*}
\mathcal{E}(s)=\mathbb{E}\left[ \left\vert
    \eta(\bX)-\eta(\bX^\prime)\right\vert
  \1_{\{(s(\bX)-s(\bX^\prime))(\eta(\bX^\prime)-\eta(\bX))<0 \}}
\right],\quad \forall s.
\end{equation*}
The type of bounds we are interested in consists in oracle inequalities in probability, which will depend on the considered family of scoring functions.
\medskip

Since $\PP$ is unknown, the minimizer $\eta$ of $L$ is unavailable. Instead, we substitute to $L$ its empirical counterpart, the empirical ranking risk $L_n$ defined as
\begin{equation*}
  L_n\colon s\mapsto \frac{1}{n(n-1)}\sum_{i\neq j}\1_{\{(Y_i-Y_j)(s(\bX_i)-s(\bX_j))<0\}},
\end{equation*}
and likewise, we let $\ern(s):=L_n(s)-L_n(\eta)$ denote the empirical excess ranking risk.
\medskip

This paper is devoted to the case where $\eta$ admits a sparse representation, \ie, only a small number $d_0\ll d$ of covariates is necessary to build efficient prediction procedures.
Note that this is also the angle studied by \cite{Li13} in a parallel work to ours, in a less general setting.

\section{PAC-Bayesian estimation of the scoring function}\label{sec:procedure}

Following \cite{GA2013}, we focus on a sparse additive modelling of the optimal scoring function. Indeed, we will build up an estimate from the family
\begin{equation*}
  \mathcal{S}_\Theta = \left\{s_\theta\colon\bx\mapsto \sum_{j=1}^d
    \sum_{k=1}^M\theta_{jk}\phi_k(x_j),\quad \theta\in\R^{dM} \right\},
\end{equation*}
where $\mathbb{D}=\{\phi_1,\dots,\phi_M\}$ is a dictionary of deterministic known functions, and we adopt the notation
$\theta = (\theta_{jk})_{j=1,\dots,d}^{k=1,\dots,M} = (\theta_{11},\theta_{12},\dots,\theta_{1M},\theta_{21},\dots,\theta_{2M},\dots,\theta_{dM}).$
Our choice for this additive formulation is motivated by the nice compromise achieved between flexibility and interpretation \citep[see for example][]{HT1986}.
Our aim is to produce a sparse estimate $\hat{\theta}$ and then compute the plugin estimator $s_{\hat{\theta}}$. To do so, we rely on the PAC-Bayesian approach and we will specify in the following section a sparsity-promoting so-called prior $\pi$ on $\Theta$ embedded with its Borel $\sigma$-algebra. Finally, let $\m = (m_1,\dots,m_d)\in\{0,1\}^d$ encode a model (where $m_j=1$ iff covariate $j$ is present).
\medskip

From the prior $\pi$, we let $\hat{\rho}_\delta$ denote the Gibbs (pseudo-)posterior density, defined as
\begin{equation}\label{gibbsposterior}
  \hat{\rho}_\delta(\mathrm{d}\theta) \propto \exp[-\delta L_n(s_\theta)]\pi(\mathrm{d}\theta),
\end{equation}
where $\delta>0$ may be seen as an inverse temperature parameter. This density twists the prior mass towards functions $s_\theta$ for which $L_{n}(s_\theta)$ is not too large. Indeed, if $\pi$ puts more probability mass on sparse vectors, $\hat{\rho}_\delta$ will favor sparse vectors with a small empirical ranking risk, thus meeting our requirements. The Gibbs pseudo-posterior in \eqref{gibbsposterior} has attracted a great deal of interest in recent years (under the name exponentially weighted aggregate as in \cite{DT2008} for example).
\medskip

The final estimator is 
\begin{equation}\label{estimateurfinal}
s_{\hat{\theta}} \colon \bX \mapsto \sum_{j=1}^d\sum_{k=1}^M\hat{\theta}_{jk}\phi_k(X_j),
\end{equation} 
where 
$\hat{\theta} \sim \hat{\rho}_{\delta}.$
Note that for the sake of brevity, we will use the notation $\hat{s}=s_{\hat{\theta}}$. The pseudo-code is summed up in \autoref{pseudocode}. We will provide in \autoref{S:theory} non-asymptotic oracle inequalities to assess the theoretical merits of the estimator $\hat{s}$. \autoref{S:mcmc} is devoted to the practical implementation of $\hat{s}$.

\begin{algorithm}[h]
\caption{Pseudo-code for our PAC-Bayesian estimator}
 \KwIn{Prior $\pi$, inverse temperature $\delta$, empirical risk $L_n$.}
 \KwOut{Estimator $s_{\hat{\theta}}$.}
 Compute 
 \begin{enumerate}
     \item Form the pseudo-posterior $\hat{\rho}_\delta(\cdot) \propto \exp[-\delta L_n(\cdot)]\pi(\cdot)$.
     \item Sample $\hat{\theta} \sim \hat{\rho}_\delta$.
     \item Compute the estimator $s_{\hat{\theta}} = \sum_{j=1}^d\sum_{k=1}^M\hat{\theta}_{jk}\phi_k$.
 \end{enumerate}
\label{pseudocode}
\end{algorithm}

\section{Oracle inequalities}\label{S:theory}

In this section, we provide the main theoretical results of the paper, consisting in oracle inequalities in probability for the estimator $\hat{s}$ defined in \eqref{estimateurfinal}. We specify different rates of convergence under several mild assumptions on the distribution $\PP$ of $(\bX,Y)$.
The only tool we need to derive our first results is an exponential inequality on the difference of the excess ranking risk and its empirical counterpart.
\medskip

\emph{For any inverse temperature parameter $\delta>0$, and any candidate
function $s$,}
\begin{equation}\label{eq:condition}
\E\exp\left[ \delta\left(\ern(s)-\er(s)\right)\right]\leq\exp(\psi(s)),
\end{equation}
\emph{where $\psi$ may depend on $n$ and $\delta$.}
\medskip

Note that this concentration condition is classical in the PAC-Bayesian literature, and allows for our first result. We let $\KL(\mu,\nu)$ denote the Kullback-Leibler divergence between two measures $\mu$ and $\nu$, and we let $\mp$ stand for the space of probability measures which are absolutely continuous with respect to $\pi$.
\begin{theo}\label{T1}
For any $\e\in(0,1)$,
\begin{multline*}
  \PP\left[\er(\hat{s})\leq \underset{\rho\in\mp}{\inf}\
    \left\{ \int \er(s)\rho(\mathrm{d}s)+\int\frac{\psi(s)}{\delta}\rho(\mathrm{d}s) + \frac{\psi(\hat{s})+2\log(2/\e)+2\KL(\rho,\pi)}{\delta} \right\}\right] \\ \geq 1-\e.
\end{multline*}
\end{theo}
This result is in the spirit of classical PAC-Bayesian bounds such as in \cite{Cat2004}. It ensures that the excess risk of our procedure $\hat{s}$ is bounded with high probability by the mean excess risk of any realization of some posterior distribution $\rho$ absolutely continuous with respect to some prior $\pi$, up to remaining terms involving the Kullback-Leibler divergence between $\rho$ and $\pi$ and the right-hand side term from the exponential inequality \eqref{eq:condition}. Note that if the right-hand term of \eqref{eq:condition} does not depend on the scoring function, \ie, $\psi(s)=\psi$ for any $s$, \autoref{T1} amounts to the inequality
\begin{equation*}
\PP\left[\er(\hat{s})\leq \underset{\rho\in\mp}{\inf}\
    \left\{ \int \er(s)\rho(\mathrm{d}s)+\frac{2\psi+2\log(2/\e)+2\KL(\rho,\pi)}{\delta} \right\}\right] \geq 1-\e.
\end{equation*}
As a matter of fact, the following lemma provides such an upper bound of the right-hand side of \eqref{eq:condition} which does not depend on the choice of $s$ \emph{and} holds for any distribution of $(\bX,Y)$.
\begin{lemma}\label{lemmacond}
For any distribution of the random variables $(\bX,Y)$, \eqref{eq:condition} holds with
$\psi(s)\equiv\delta^2/4n$.  
\end{lemma}
\begin{coro}\label{coro1}
Choosing $\delta=\sqrt{n}$ in \autoref{T1}, for any $\e\in(0,1)$,
\begin{equation*}
  \PP\left[\er(\hat{s})\leq \underset{\rho\in\mp}{\inf}\
    \left\{ \int \er(s)\rho(\mathrm{d}s) + \frac{1/2+2\log(2/\e)+2\KL(\rho,\pi)}{\sqrt{n}} \right\}\right]\geq 1-\e.
\end{equation*}
\end{coro}

The message here is that we obtain the classical slow rate of convergence $\sqrt{n}$ (as achieved, for example, by empirical AUC minimization on a Vapnik-Cervonenkis class) under no assumption whatsoever on $\PP$ with the PAC-Bayesian approach.

\subsection{Using a sparsity-promoting prior}

Our goal is to obtain \emph{sparse} vectors $\theta$, and this constraint is met with the introduction of the following prior $\pi$. For any compact set $A$, let $\mathrm{Unif}_A$ stand for the uniform distribution on $A$ given by
\begin{equation*}
\mathrm{Unif}_A(B) = \frac{\mu(B)}{\mu(A)}, \quad B\subseteq A,
\end{equation*}
where $\mu$ is the reference Lebesgue measure.
In addition we denote by $|\m|_0=\sum_{j=1}^d m_j$ the $\ell^0$ norm of $\m$. We let $\mathcal{B}_\m$ denotes the $\ell^2$-ball in $\R^{|\m|_0}$ of radius $2$. We then define the sparsity-inducing prior as
\begin{equation}\label{eq:prior}
  \pi(\mathrm{d}\theta)\propto\sum_{\m}
  \binom{d}{|\m|_0}^{-1}\beta^{|\m|_0 M}\mathrm{Unif}_{\mathcal{B}_\m}(\theta),
\end{equation}
where $\beta\in (0,1)$.
This prior may be traced back to \cite{LB2006} and serves our purpose: for any $\theta\in\Theta$, its probability mass will be negligible unless its support has a very small dimension, \ie, $\theta$ is sparse. Next, we introduce a technical condition required in our scheme.

\begin{cond}\label{cond_dens} There exists c>0, such that
$$\mathbb{P}[s_{\theta}(\bX)-s_{\theta}(\bX') \geq 0, s_{\theta'}(\bX)-s_{\theta'}(\bX')\leq 0]\leq c\|\theta-\theta'\|
$$
for any $\theta$ and $\theta'\in\mathbb{R}^{d}$  such that $\|\theta\|=\|\theta'\|=1$.
\end{cond}

This assumption is the exact analogous to the density assumption used in \cite{RACL2014} and echoes classical technical requirements linked to margin assumptions, as discussed further \citep[see for example][]{AudTsy07}.
\medskip

The use of this sparsity-inducing prior allows us to obtain terms in the right-hand side of the oracle inequalities which depend on the intrinsic dimension of the ranking problem, \ie, the dimension of the sparsest representation $s_\theta$ of the optimal scoring function.

\begin{theo}\label{coro1KL}
Let $\delta=\sqrt{n}$ and let \autoref{cond_dens} hold.
  With the prior $\pi$ defined as in \eqref{eq:prior},
  we obtain for any $\e\in(0,1)$,
  \begin{equation*}
    \PP\left[
    \er(\hat{s})\leq \underset{\m}{\inf}\ \underset{\theta\in\mathcal{B}_\m, \|\theta\|=1}{\inf}\
      \left\{ 
      \er(s_\theta)  + \frac{ 3/2+2\log(2/\e)+ \log(2c\sqrt{n})+K}{\sqrt{n}}
      \right\}
      \right] 
      \geq 1-\e,
  \end{equation*}
  where
    $K = 2\left(|\m|_0M\log(1/\beta)+|\m|_0\log\frac{de}{|\m|_0}+\log\frac{1}{1-\beta}\right).$
\end{theo}

This sharp oracle inequality ensures that if there exists indeed a sparse representation (\ie, some sparse model $\m$) of the optimal scoring function, \ie, involving only a small number of covariates, then the excess risk of our procedure is bounded by the best excess risk among all linear combinations of the dictionary up to some small terms. On the contrary, if such a representation does not exist, $|\m|_0$ is comparable to $d$ and terms  like $|\m|_0\log(d)/\sqrt{n}$ and $|\m|_0 M\log(1/\beta)$ start to emerge.

\subsection{Faster rates with a margin condition}

In order to obtain faster rates, we follow \cite{REJS} and work under the following margin condition.

\begin{cond}\label{cond-margin}
The distribution of $(\bX,Y)$ verifies the margin assumption {\bf MA($\alpha$)} with parameter $0\leq\alpha\leq 1$ if there exists $C<\infty$ such that:
\begin{equation*}
\PP\left[(s(\bX)-s(\bX^\prime))(\eta(\bX)-\eta(\bX^\prime))<0\right] \leq C (L(s)-L^\star)^{\frac{\alpha}{1+\alpha}},
\end{equation*}
for any scoring function $s$.
\end{cond}
This margin (or low noise) condition was first introduced for classification by \cite{MT1999}, 
later adapted to the ranking problem by \cite{CLV08}.
Note that this statement is trivial for the value $\alpha=0$ and increasingly restrictive as $\alpha$ grows. We refer the reader to \cite{BBL2005}, \cite{Lecue06} and \cite{REJS} for an extended discussion.

%

Finally, for the sake of brevity, we will use the notation $\Upsilon$, $C_{1}$, $C_{2}$ and $C_3$ for generic constants in the following statements. The exact form of those constants may be found in the proofs (\autoref{S:proof}).

\begin{theo}\label{coro2}
Assume that \autoref{cond-margin} holds and let $\delta=\Upsilon n^{\frac{1+\alpha}{2+\alpha}}$. For any $\e\in(0,1)$,
\begin{multline*}
 \PP\left[\er(\hat{s})\leq \underset{\rho\in\mp}{\inf}\
    \left\{ 3\int \er(s)\rho(\mathrm{d}s) +
      n^{-\frac{1+\alpha}{2+\alpha}}\left[C_1+\Upsilon^{-1}\left(1/2+\log(2/\e)+\KL(\rho,\pi)\right)\right]
       \right\}\right] \\ \geq 1-\e,
\end{multline*}
where $\Upsilon$ and $C_1$ are constants depending only on $c$, $C$, $\alpha$ and $\C$.
\end{theo}

Note that taking $\alpha=0$ yields a rate of convergence similar to the one in \autoref{coro1}. A trade-off is at work in that result: in all generality, the fastest rate achievable is of magnitude $\sqrt{n}$. However, under the margin condition, refined rates are available, at the cost of generality: the greater $\alpha$, the faster the rate \emph{and} the more restrictive the assumption on $\PP$. For more comments on the introduction of margin conditions for ranking problems and its impact on rates of convergence, we refer the reader to the aforecited \cite[Section 2.3]{CRICML}.
\medskip

The next result is the adaptation of \autoref{coro1KL} under \autoref{cond-margin}, to obtain faster rates.

\begin{theo}\label{KL}
Assume that \autoref{cond_dens} and \autoref{cond-margin} holds and let $\delta=\Upsilon n^{\frac{1+\alpha}{2+\alpha}}$. With the prior $\pi$ defined as in \eqref{eq:prior}, we obtain for any $\e\in(0,1)$,
  \begin{multline*}
    \PP\left[\er(\hat{s})\leq \underset{\m}{\inf}\ \underset{\theta\in\mathcal{B}_\m,\|\theta\|=1}{\inf}\
    \left\{ 3\er(s_\theta)
      +n^{-\frac{1+\alpha}{2+\alpha}}C_1 \left(K+3/2+2\log(2/\e)+\log\left(n^{\frac{1+\alpha}{2+\alpha}}\right)\right)
    \right\}\right] \\ \geq 1-\e,
  \end{multline*}
where $\Upsilon$ and $C_1$ are constants depending only on $c$, $C$, $\alpha$, $\beta$ and
$\C$, and
  $K = 2\left(|\m|_0M\log(1/\beta)+|\m|_0\log\frac{de}{|\m|_0}+\log\frac{1}{1-\beta}\right).$
\end{theo}
This result walks in the footsteps of previous works on the use of the margin condition in bipartite ranking, such as in \cite{REJS}.
As in \autoref{coro1KL}, \autoref{KL} exhibits right-hand terms in the oracle inequality which depend on the intrinsic dimension of the problem, now with a significantly faster rate, of magnitude $n^{\frac{1+\alpha}{2+\alpha}}$.

\subsection{Rates of convergence on Sobolev classes}

In order to control the bias leading term in the previous oracle inequalities, we refine the previous results under the additional assumption that the regression function now belongs to some functional regularity space. Following \cite{Tsy2009}, we consider the Sobolev ellipsoid defined as
\begin{equation*}
\mathcal{W}(\tau,\kappa) = \left\{f\in \mathrm{L}^2([-1,1])\colon
    f=\sum_{k=1}^\infty\theta_{k}\varphi_k \quad\mathrm{and}\quad
    \sum_{i=1}^\infty i^{2\tau}\theta_{i}^2\leq \kappa \right\}.
\end{equation*}

\begin{cond}\label{condSOB}
$\eta=\sum_{j\in S^\star}\eta_j$, and for all $j=1,\dots,d$, $\eta_j\in\mathcal{W}(\tau,\kappa)$.
\end{cond}
In other words, we now assume that a sparse (additive) representation of $\eta$ does exist with some sufficient regularity, and that its support is some ensemble $S^\star\subset \{1,\dots,d\}$.
\medskip

We are now in a position to state our next result, which is again an adaptation of \autoref{coro1KL}.

\begin{theo}\label{Sobolev}
Assume that \autoref{cond_dens} and \autoref{condSOB} hold, and let $\delta= \Upsilon n^{\frac{1+\tau}{1+2\tau}}$. With the prior $\pi$ defined as in \eqref{eq:prior} we obtain for any $\e\in(0,1)$,
    \begin{equation*}
	\PP\left[\er(\hat{s})\leq \left\{
C_1 n^{-\frac{\tau}{2\tau+1}} +C_2 \left(2\log(2/\e)+K+|S^\star|_0\log(C_{3}n^{\frac{\tau+1}{2\tau+1}})\right) n^{-\frac{\tau+1}{2\tau+1}} \right\}\right]\geq 1-\e,
\end{equation*}
where $\Upsilon$, $C_1$, $C_2$ and $C_3$ are constants depending only on $c$, $C$, $\beta$, $\kappa$, $|S^\star|_0$ and $\tau$, and $K =2\left(|S^\star|_0\log\frac{de}{|S^\star|_0}+\log\frac{1}{1-\beta}\right).$
\end{theo}
The leading term $\mathcal{O}\left(n^{-\frac{\tau}{2\tau+1}}\right)$ is the classical nonparametric rate of convergence for the estimation of a function with some regularity $\tau$, and the other terms involve the dimension of the best approaching model: in that sense, if a sparse representation of $\eta$ exists, these terms will be small. Note that this result proves that our estimator $\hat{s}$ is adaptive to the unknown sparsity pattern $S^\star$.
\medskip

Our last oracle inequality is our most detailed result, and combines the settings of \autoref{KL} and \autoref{Sobolev}.

\begin{theo}\label{Sobolev-MA}
Assume that \autoref{cond_dens}, \autoref{cond-margin} and \autoref{condSOB} hold. Let $\delta=\Upsilon n^{\frac{1+\tau(1+\alpha)}{1+\tau(2+\alpha)}}$. With the prior $\pi$ defined as in \eqref{eq:prior}, we have for any $\e\in(0,1)$,
 \begin{multline*}
	\PP\left[\er(\hat{s})\leq \left\{
C_1 n^{\frac{-\tau(1+\alpha)}{1+(2+\alpha)\tau}}+C_2 \left(2\log(2/\e)+K +|S^\star|_0\log\left(C_3 n^{\frac{1+\tau(1+\alpha)}{1+\tau(2+\alpha)}}\right)\right) n^{\frac{-(1+\tau(1+\alpha))}{1+(2+\alpha)\tau}}
 \right\}\right] \\ \geq 1-\e,
\end{multline*}
where $\Upsilon$, $C_1$, $C_2$ and $C_3$ are constants depending only on $c$, $C$, $\beta$, $\kappa$, $|S^\star|_0$, $\tau$, $\mathcal{C}$ and $\alpha$, and $K =2\left(|S^\star|_0\log\frac{de}{|S^\star|_0}+\log\frac{1}{1-\beta}\right).$
\end{theo}

Again, note that this result proves our estimator $\hat{s}$ to be fully adaptive to the unknown sparsity pattern $S^\star$. However let us mention that since $\delta$ depends on the smoothness $\tau$ and to the margin parameter $\alpha$, our estimator is not adaptive to those parameters. Yet it is possible to make our procedure adaptive by using the strategy proposed by \cite{REJS} consisting in an aggregation procedure to combine estimators with exponential weights, over a grid for both $\alpha$ and $\tau$.
\medskip

Let us conclude this section by a comment on the links between our results and the minimax results introduced in \cite{CRICML}. To the best of our knowledge, \cite{CRICML} are the first to prove minimax optimality results for the bipartite ranking problem, with oracle inequalities in expectation. The minimax rate of convergence is exactly the one appearing in \autoref{Sobolev-MA}, namely $\mathcal{O}\left(n^{\frac{-(1+\tau(1+\alpha))}{1+(2+\alpha)\tau}}\right).$
Since our results hold in probability, it is straightforward to obtain the similar oracle inequalities in expectation (integrating with respect to $\e$). Our PAC-Bayesian estimator thus achieves the minimax rate of convergence for the bipartite ranking problem, moreover on a larger functional class (Sobolev ellipsoid vs. H\"older class).

\section{MCMC implementation}\label{S:mcmc}

Our approach requires to sample from the Gibbs posterior $\hat{\rho}_{\delta}$ and we rely on an MCMC procedure to do so. However, several pitfalls appear: since $\hat{\rho}_{\delta}$ is a distribution on a very high dimensional space with a complex structure, classical MCMC algorithms are likely to perform poorly. We therefore propose an adaptation to the ranking setting of the algorithm presented in \cite{GA2013} in the context of regression, which is inspired by \cite{PD2012}. The key idea lies in the definition of a neighborhood relationship among the different models. Let us recall that $\m = (m_1,\dots,m_d)\in\{0,1\}^d$ denotes a model. Our transdimensional gateway is defined as follows: at each MCMC step, we propose to add a missing covariate to the current model, delete an existing one, or keep the same covariates. This entices the definition of three possible neighborhoods for a model $\m^{t}$:
\begin{itemize}
\item The set of all models having the same covariates that $\m^{t}$, plus one, denoted $\mathcal{V}^{+}_{\mathbf{m}_t}$,
\item The set of all models having the same covariates that $\m^{t}$, minus one, denoted $\mathcal{V}^{-}_{\mathbf{m}_t}$,
\item The neighborhood corresponding to the case where no dimension change is proposed at iteration $t$, which is limited to the current model $\m^{t}$.
\end{itemize}
We first select a neighborhood (\ie, a move) among $\mathcal{V}^{+}_{\mathbf{m}_t}$, $\mathcal{V}^{-}_{\mathbf{m}_t}$ and $\{\m^{t}\}$ with probabilities $(a,a,b)$ (\emph{e.g.}, $a=1/4$ and $b=1/2$ in the following). Let $\mathcal{V}$ denote the selected neighborhood. For any model in $\mathcal{V}$, a candidate vector $\theta$ is sampled from a proposal Gaussian distribution, whose mean is a benchmark estimator (such as least-squares fit, the maximum likelihood estimator, a Lasso estimator, etc.) and whose variance is a parameter to the algorithm. The joint move towards the candidate model and estimator is then accepted following a Metropolis-Hastings ratio. This approach has been implemented for the regression problem in \cite{Gue2013a}.
\medskip

Note that this algorithm is designed to sample from $\hat{\rho}_{\delta}$. However, estimators sampled from this algorithm may suffer from large variance, and we introduce a more stable version of our algorithm. Recall that in \eqref{estimateurfinal}, $\hat{\theta}$ is sampled from $\hat{\rho}_{\delta}$. From a numerical stability perspective, it is useful to consider the mean of $\hat{\rho}_{\delta}$ instead. Thus we adapt our notation and now define two different estimators: $s_{\hat{\theta}^{r}}$ (same estimator as in \eqref{estimateurfinal}) and $s_{\hat{\theta}^{a}}$, where
\begin{equation}\label{estimateurrand}
\hat{\theta}^{r} \sim \hat{\rho}_{\delta}\quad\textrm{(randomized estimator)},
\end{equation}
and
\begin{equation}\label{estimateuragg}
\hat{\theta}^{a} = \int_\Theta \theta\hat{\rho}_\delta(\mathrm{d}\theta) 
= \mathbb{E}_{\hat{\rho}_\delta}\theta \quad\textrm{(averaged estimator)}.
\end{equation}
In the following results, estimators defined in \eqref{estimateurrand} and \eqref{estimateuragg} will be referred to as \emph{PAC-Bayesian Randomized} and \emph{PAC-Bayesian Averaged}, respectively.
Finally, we introduce the following notation: for any model $\m$, we let $\theta_{\m}$ denote a benchmark estimator, $\varphi_{\m}$ denotes the density of the Gaussian distribution $\mathcal{N}(\theta_{\m},\sigma^{2}I_{\m})$ where $I_{\m}$ stands for the identity matrix $|\m|_{0}\times |\m|_{0}$.

The pseudocode is presented in \autoref{algo:pac_guedj}.
\medskip

\begin{algorithm}[h!]
\caption{An MCMC algorithm for PAC-Bayesian estimators}
\raggedright \textbf{Input}: 
\begin{tabular}{l}
horizon $T$, \\
burnin $b$, \\
proposal variance $\sigma^2>0$, \\
inverse temperature parameter $\delta >0$.
\end{tabular}

\raggedright \textbf{Output}: two sequences of models $(\mathbf{m}^t)_{t=1}^T$ and estimators $(\theta^t)_{t=1}^T$.
\medskip

\raggedright At time $t=2,\dots,T$, 
\begin{description}
\item [{1:}] Pick a move and form the corresponding neighborhood $\mathcal{V}_{t}$.
\item [{2:}] For all $\mathbf{m}\in\mathcal{V}_{t}$, draw a candidate estimator $\tilde{\theta}_{\m}\sim\mathcal{N}(\theta_{\m},\sigma^2 I_{\m})$.
\item [{3:}] Pick a pair $(\mathbf{m},\tilde{\theta}_{\m})$ with probability proportional to $\hat{\rho}_\delta(\tilde{\theta}_{\m})/\varphi_{\m}(\tilde{\theta}_{\m})$.
\item [{4:}] Set \begin{equation*}
\begin{cases}
\theta^t=\tilde{\theta}_{\m} \\
\mathbf{m}^t=\mathbf{m}
\end{cases} \textrm{with probability } \alpha, \qquad
\begin{cases}
\theta^t= \theta^{t-1}\\
\mathbf{m}^t=\m^{t-1}
\end{cases} \textrm{with probability } 1-\alpha,
\end{equation*}
$$\alpha=\min\left(1,\frac{\hat{\rho}_\delta(\tilde{\theta}_{\m})\varphi_{\m}(\theta^{t-1})}{\hat{\rho}_\delta(\theta^{t-1})\varphi_{\m}(\tilde{\theta}_{\m})}\right).$$
\end{description}
\raggedright \textbf{Final estimators}:
\begin{align*}
s_{\hat{\theta}^{r}}  \colon \bX \mapsto \sum_{j=1}^d\sum_{k=1}^M \theta^T_{jk}\phi_k(X_j) \qquad &\textrm{(PAC-Bayesian Randomized)}, \\
s_{\hat{\theta}^{a}} \colon \bX \mapsto \sum_{j=1}^d\sum_{k=1}^M\left(\sum_{\ell=b+1}^T \theta^\ell_{jk}\right)\phi_k(X_j) \qquad &\textrm{(PAC-Bayesian Averaged)}.
\end{align*}
\label{algo:pac_guedj}
\end{algorithm}

Recall that our procedure relies on the dictionary $\mathbb{D}$. In our implementation, we chose $M=13$ (a value achieving a compromise between computational feasibility and analytical flexibility) and as functions the seven first Legendre polynomials and the six first trigonometric polynomials. Let us stress here that $M$ characterizes the flexibility of the representation of $\mathcal{S}_\Theta$. The larger the better, yet in practice we have found that small values were largely enough to capture complex phenomena. Linear scoring functions correspond to the simplest case $M=1$.
\medskip

The algorithm mainly depends on two input parameters, the variance of the proposal distributions $\sigma^{2}$ and the inverse temperature parameter $\delta>0$. Bad choices for these two parameters are likely to quickly deteriorates the performance of the algorithm. Indeed, if $\sigma^{2}$ is large, the proposal gaussian distribution will generate candidate estimators weakly related to the benchmark ones. On the contrary, small values for $\sigma^{2}$ will concentrate candidates towards benchmark estimators, whittling the diversity of estimators proposed. It is important to note that proposing randomized candidate estimators is a key part of our work. We resort to cross-validation to select $\sigma^2$.
\medskip

As for the inverse temperature parameter $\delta$, let us recall here that even though optimal values are proposed for $\delta$ in each theorem in \autoref{S:theory}, this is of little comfort in practice as those values typically depend on unknown quantities such as $\alpha$ or $\tau$. Therefore we again resort to cross-validation to select $\delta$. This choice is sensitive to the performance of \autoref{algo:pac_guedj}. Indeed, small values clearly make the Gibbs posterior very similar to the sparsity-inducing prior. In that case, fit to the data is negligible, and the performance are likely to drop. On the contrary, large values for $\delta$ will concentrate most of the mass of the Gibbs posterior towards a minimizer of the empirical ranking risk, which is possibly non-sparse.
\medskip

\begin{figure}[h]
  \caption{Numerical experiments on synthetic data. The two meaningful covariates are the third and the fifth.}
\label{figsynthetic}
  \subfloat[{Boxplot of AUC for both estimators. Blue dotted line: optimal oracle value (.7387).}]{\includegraphics[width=.49\textwidth]{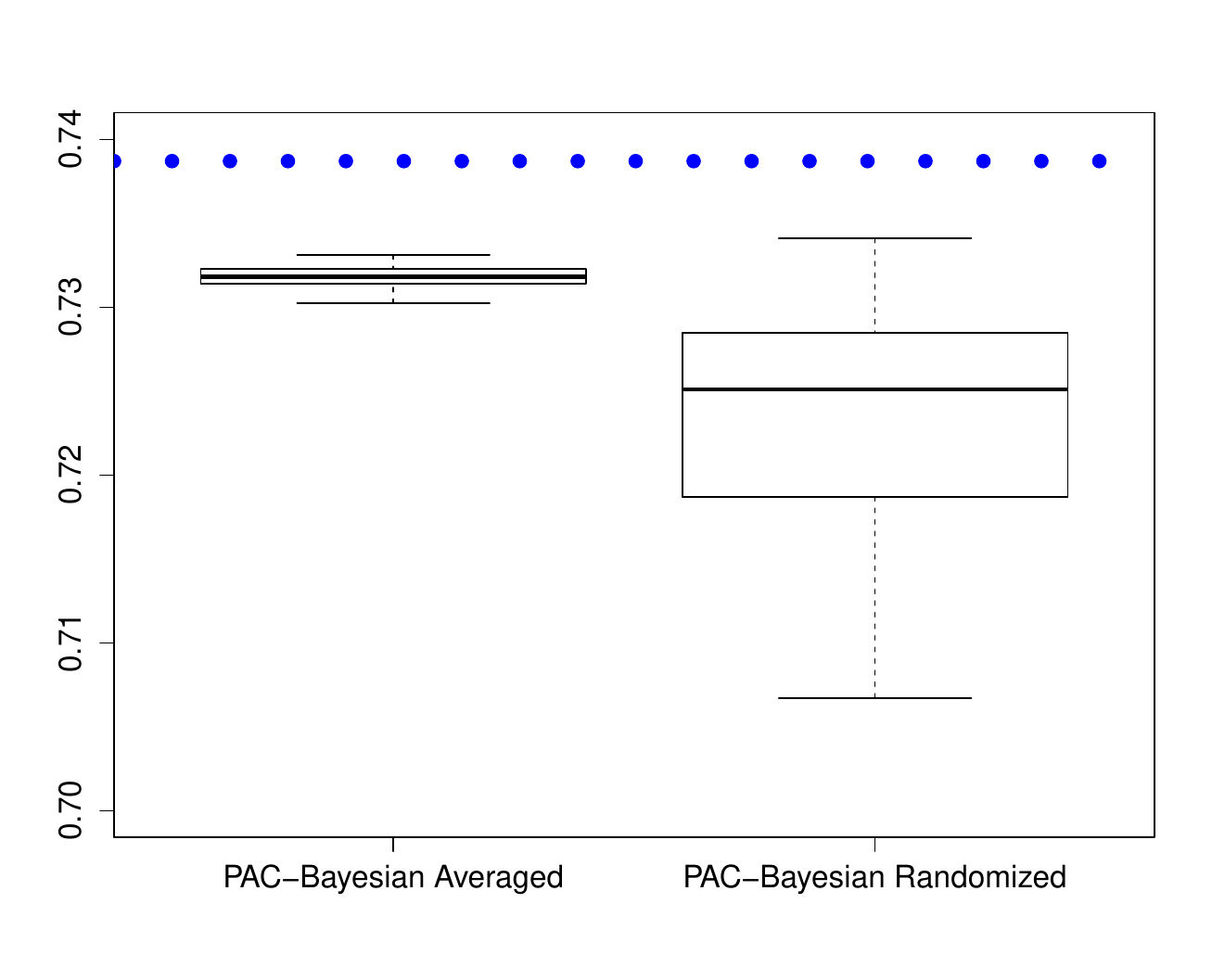}}
  \hfill
  \subfloat[Selected variables along the MCMC chain.]{\includegraphics[width=.49\textwidth]{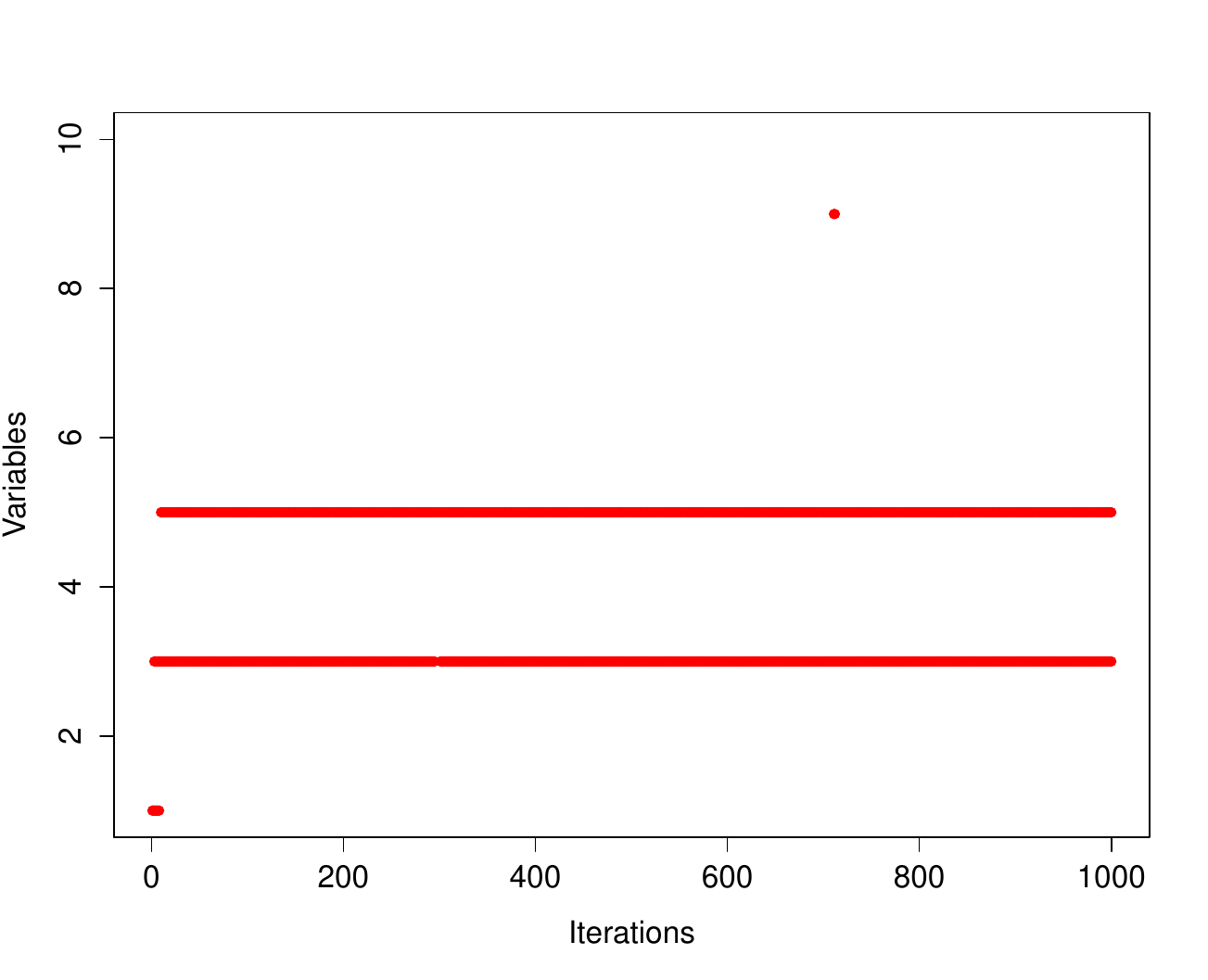}} \\
  \hfill
  \subfloat[For the third covariate, green points are $\eta(X_{i})$ and red ones are $s_{\hat{\theta}^{a}} (X_i)$.]{\includegraphics[width=.49\textwidth]{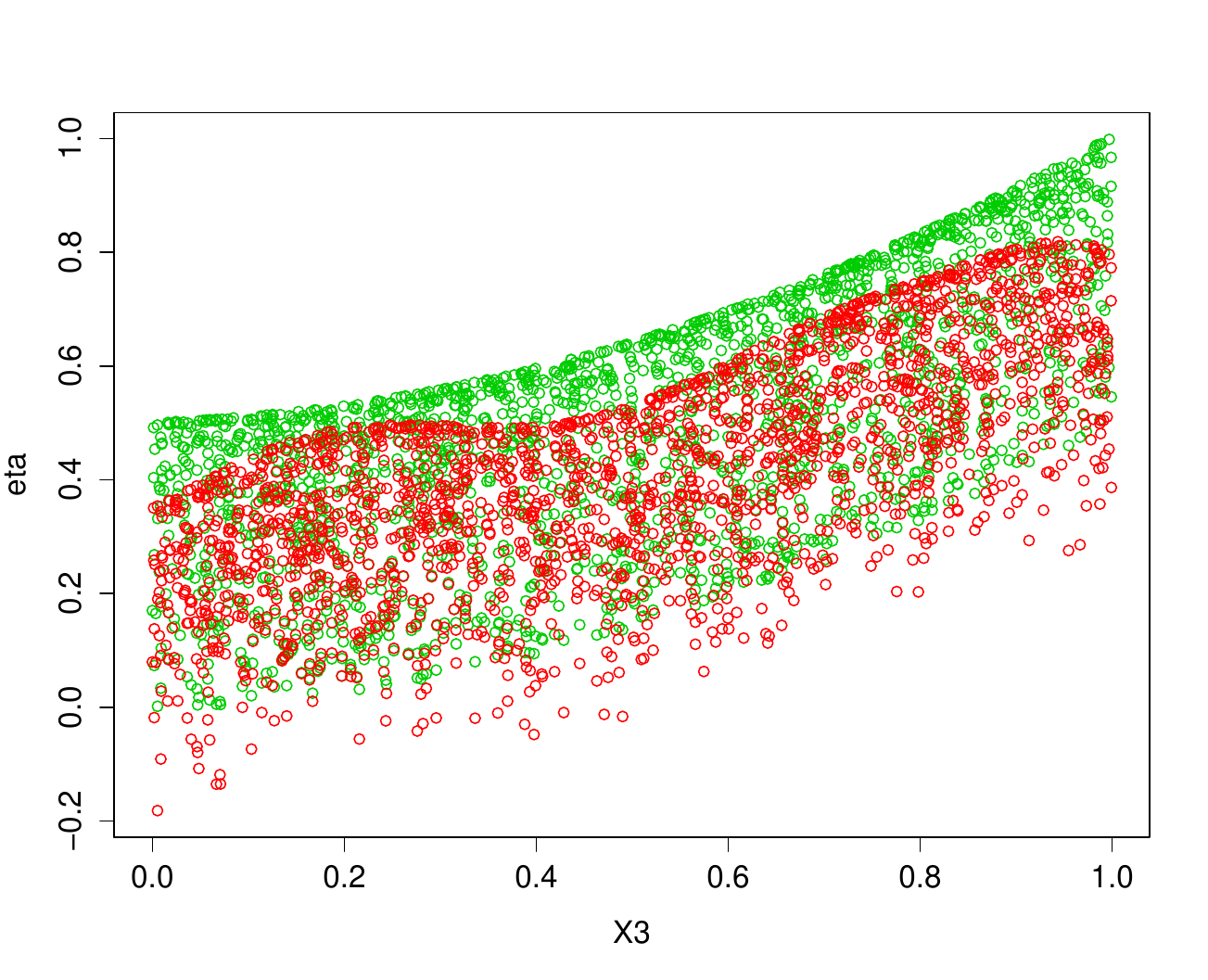}}
  \hfill
  \subfloat[For the fifth covariate, green points are $\eta(X_{i})$ and red ones are $s_{\hat{\theta}^{a}} (X_i)$.]{\includegraphics[width=.49\textwidth]{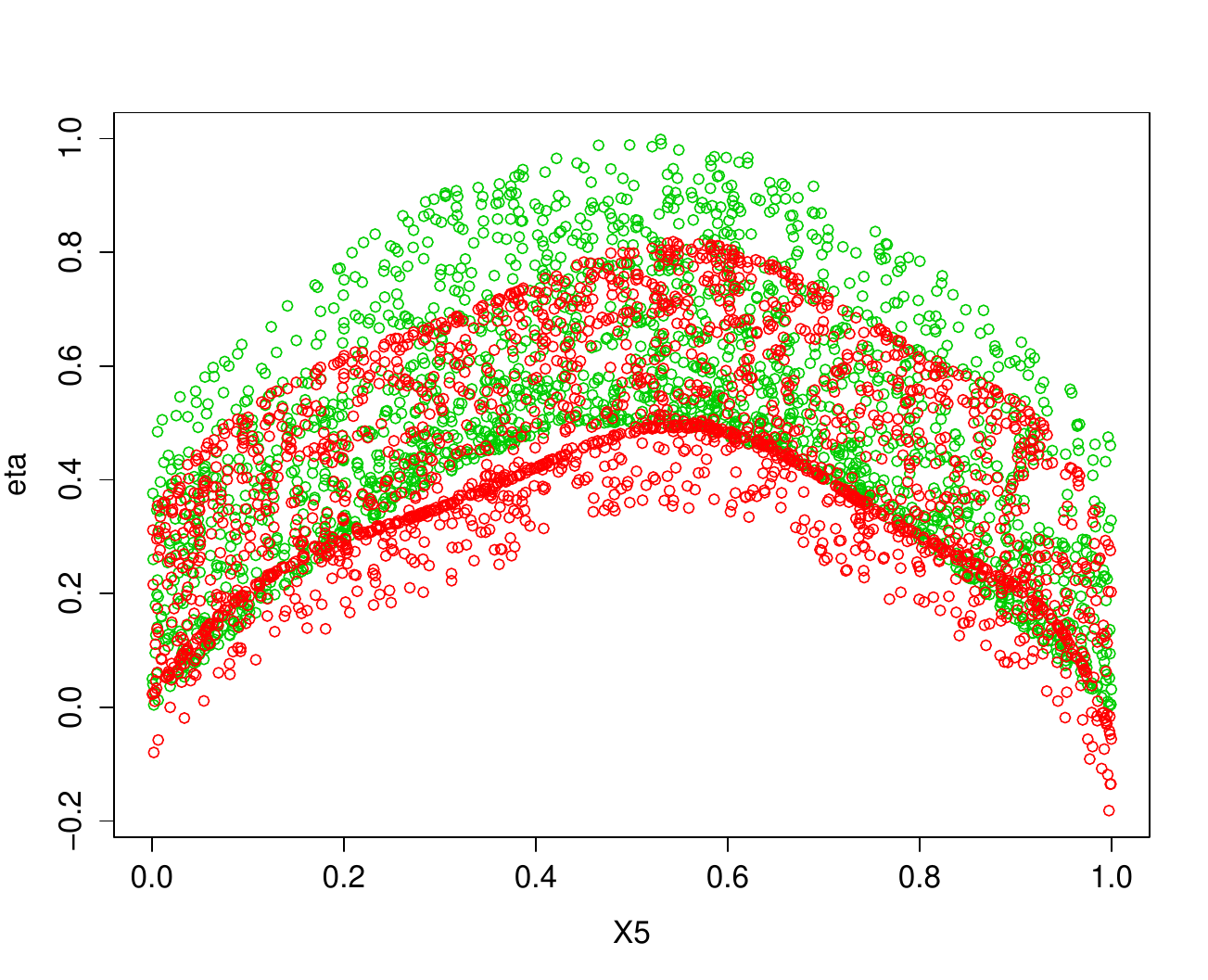}}
\end{figure}

We illustrate the behavior of \autoref{algo:pac_guedj} with several values for the two parameters $\sigma^{2}$ and $\delta$, on the following synthetic model.
\begin{multline}\label{modele}
n_{\textrm{train}}=1000,\quad n_{\textrm{test}}=2000,\quad d=10,\quad \bX\sim\mathcal{U}([0,1]^d),\\ Y = \1\{\eta(\bX)>\mathcal{U}([0,1])\},\quad  \eta\colon \bx \mapsto X_3 + 7X_3^2 + 8\sin(\pi X_5)
\end{multline}
Note that the values taken by $\eta$ have been renormalized to fit in $(0,1)$. We take as a benchmark in our simulations the AUC of $\eta$ as defined in \eqref{modele}, which is .7387. In other words, the closer to .7387, the better the performance. Our simulations are summed up in \autoref{tablesynthetic} and \autoref{figsynthetic}.

\begin{table}[h]
\caption{Mean (and variance) of AUC over 50 replications for both estimators \eqref{estimateurrand} and \eqref{estimateuragg} (1000 MCMC iterations, 800 burnin iterations), and frequencies of the selected variables for the last 200 iterations.}
\label{tablesynthetic}
\centering
\small{
\begin{tabular}{ccccccccc}
  \hline
  $\delta$ & $\sigma^2$ & PAC-Bayesian & PAC-Bayesian & $\# 3$ & $\# 5$ & $\sum_{i \neq \{3,5\}}\# i$ \\ 
  & & Averaged & Randomized & & & \\
  \hline
 100 & 1& .699  (.020) & .697 (.019) & .9363 & .9200 & .0272 \\ 
  100 & .1& .713 (.014) & .712  (.014)  & .9881 & .9800 & .0369 \\ 
  100 & .01 & .723  (.009) & .721  (.010)  & 1.0000 & 1.0000 & .0617 \\ 
   100 & .001 & .720  (.006) & .712 (.006)  & 1.0000 & 1.0000 & .4251 \\ 
   10 & 1& .701  (.024) & .702  (.024)  & .9585 & .9000 & .0396 \\ 
   10 & .1 & .713  (.012) & .712  (.014)  & 1.0000 & 1.0000 & .0561 \\ 
   10 & .01 & .724  (.008) & .724  (.008)  & 1.0000 & 1.0000 & .0404 \\ 
   10 & .001 & .721  (.006) & .719  (.006)  & 1.0000 & 1.0000 & .3970 \\ 
   1 & 1 & .705  (.019) & .705 (.014)  & 1.0000 & 1.0000 & .0898 \\ 
  1 & .1 & .715  (.012) & .714 (.012)  & 1.0000 & 1.0000 & .0672 \\ 
   1 & .01 & .725  (.004) & .725  (.004) & 1.0000 & 1.0000 & .0286 \\ 
   1 & .001& .727  (.003) & .725  (.004)  & 1.0000 & 1.0000 & .1443 \\ 
   .1 & 1 & .703  (.014) & .700  (.014)  & 1.0000 & 1.0000 & .0977 \\ 
   .1 & .1 & .716  (.010) & .709  (.013)  & .9996 & .9981 & .0304 \\ 
   .1 & .01 & .729 (.003) & .717  (.010) & .9998 & .9982 & .0069 \\ 
   .1 & .001 & .731  (.001)& .723  (.006)  & .9995 & .9996 & .0004 \\ 
   .01 & 1 & .663  (.045) & .563  (.058)  & .5323 & .4641 & .1913 \\ 
  .01 & .1 & .667  (.058) & .561  (.068)  & .3811 & .4426 & .0620 \\ 
   .01 & .01 & .672  (.017) & .634  (.028)  & .4293 & .5783 & .0059 \\ 
   .01 & .001 & .663 (.019) & .646  (.042)  & .5743 & .3700 & .0089 \\ 
   \hline
\end{tabular}
}
\end{table}

\paragraph{Comments} The overall performance is good, and the best AUC value (equal to .731, achieved by the averaged estimator with $\delta=.1$ and $\sigma^{2}=.001$) is very close to the optimal oracle value .7387, therefore validating our procedure.

As expected, the averaged estimator exhibits slightly better performance (see \autoref{figsynthetic}, a), due to its improved numerical stability over the randomized estimator.
\medskip

When $\delta$ and $\sigma^{2}$ are finely calibrated, the algorithm almost always selects the two ground truth covariates ($3$ and $5$, see \autoref{figsynthetic}, b). As highlighted above, additional junk covariates may be selected when the fit to the data is weak (\ie, $\delta$ is too small), leading to poor performance.

We now compare our PAC-Bayesian procedure to two state-of-the-art methods, on real-life datasets. Since our work investigates nonlinear scoring functions, we restricted the comparison with similar methods. The Rankboost \citep{FISS03} and TreeRank \citep{BCDV2010,CDV2011} algorithms appear as ideal benchmarks.
We have conducted a series of experiments on the following datasets: Diabetes, Heart, Iono, Messidor, Pima and Spectf. All these datasets are freely available online, following \href{http://archive.ics.uci.edu/ml/}{http://archive.ics.uci.edu/ml/} and serve as classical benchmark for machine learning tasks. Our results are wrapped up in \autoref{tablereal} (where TRT, TRl and TRg denote the TreeRank algorithm trained with decision trees, linear SVM and gaussian SVM, respectively). For most datasets, our PAC-Bayesian estimators compete on similar ground with the four other methods, intercalating between the less and most performant method, while being the only ones supported by ground theoretical results.

\begin{table}[h]
\caption{Cross-validated mean (and variance) of AUC over seven real-life datasets.}
\scriptsize{
\begin{tabular}{lcccccc}
  \hline
 Name ($n$, $d$) & PAC-B. & PAC-B. & TRT & TRl & TRg & Rankboost \\ 
 & Averaged & Randomized & & & & \\
  \hline
Vowel (990,10) &  .848 (.018) & .846 (.017) & .908 (.017) & .946 (.011) & .976 (.009) & .946 (.013) \\ 
Messidor (1151,11) & .738 (.033) & .732 (.034) & .687 (.038) & .800 (.023) & .754 (.036) & .747 (.021) \\ 
Iono (351,32) & .871 (.047) & .869 (.047) & .878 (.033) & .846 (.050) & .905 (.024) & .929 (.017) \\ 
Diabetes (768,8) & .772 (.038) & .767 (.040) & .777   (.037) & .810 (.036) & .794 (.037) & .820 (.033) \\ 
Heart (270,5) & .733 (.061) & .725 (.065) & .700 (.072) & .752 (.063) & .676 (.077) & .746 (.062) \\ 
Pima (768,8) & .782 (.036) & .772 (.038) & .777 (.037) & .810 (.036) & .703 (.033) & .820 (.033) \\ 
Spectf (267,44) & .764 (.103) & .759 (.102) & .757 (.129) & .711 (.109) & .601 (.089) & .854 (.111) \\ 
   \hline
\end{tabular}
}
\label{tablereal}
\end{table}

\section{Conclusion}

We study in the present paper the problem of bipartite ranking in its theoretical and algorithmic aspects, in a high-dimensional setting through the PAC-Bayesian approach. Our model is nonlinear and assumes a sparse additive representation of the optimal scoring function. We propose an estimator based on the Gibbs pseudo-posterior distribution, and derive oracle inequalities in probability under the sparsity assumption, \ie, with terms involving the intrinsic dimension instead of the ambient dimension $d$. Under minimal assumption, we recover classical rates of convergence $\mathcal{O}(n^{-1/2})$. On a Sobolev ellipsoid with regularity $\tau$ and under a margin assumption of parameter $\alpha$, we obtain the minimax rate of convergence $\mathcal{O}\left(n^{\frac{-(1+\tau(1+\alpha))}{1+(2+\alpha)\tau}}\right).$ A salient fact is that our results significantly extend previous works \citep{CRICML}, in that our inequalities hold in probability (instead of inequalities in expectation). A summary of all our theoretical claims is presented in \autoref{tablesummary}.
\medskip

\begin{table}[h]
\caption{Summary of the theoretical results.}
\label{tablesummary}
\begin{center}
\scriptsize{
\begin{tabular}{ccccl}
    Result & Ass. \ref{cond_dens} & Ass. \ref{cond-margin} & Ass. \ref{condSOB} & Comment \\
  \hline
  \autoref{T1} & & & & Most generic result\\
  \autoref{coro1KL} & \cmark & & & With a sparsity-inducing prior\\
  \autoref{coro2} & & \cmark & & Faster rates with margin assumption, generic prior\\
  \autoref{KL} & \cmark & \cmark & & Faster rates with margin assumption, sparsity-inducing prior \\
  \autoref{Sobolev} & \cmark & & \cmark & On Sobolev spaces, no margin assumption \\
  \autoref{Sobolev-MA} & \cmark & \cmark & \cmark & On Sobolev spaces, with a margin assumption
\end{tabular}
}
\end{center}
\end{table}

Next, we propose an implementation of the PAC-Bayesian estimators through a transdimensional MCMC. Its performance on synthetic and real-life datasets competes with other nonlinear ranking algorithms. In conclusion, the main contributions of this paper are a nonlinear procedure with provable minimax rates and an efficient implemented approximation for the bipartite ranking problem.


\label{S:conclusion}

\section{Proofs}\label{S:proof}
The following lemma \citep[Legendre transform of the
Kullback-Leibler divergence,][]{Csi1975} is a key ingredient in our proofs and the demonstration may be found in \cite[Equation
5.2.1]{Cat2004}.
\begin{lemma}\label{lemma:catoni}
  Let $(A,\mathcal{A})$ be a measurable space. For any probability $\mu$ on $(A,\mathcal{A})$ and any measurable function
  $h : A \to \R$ such that $\int(\exp\circ\, h) \rm{d}\mu < \infty$,
  \begin{equation*}
    \log\int(\exp\circ\, h) \mathrm{d}\mu = \underset{m\in\MP(A,\mathcal{A})}{\sup}
    \left\{\int h \mathrm{d}m - \KL(m,\mu)\right\},
  \end{equation*}
  with the convention
  $\infty-\infty =-\infty$. Moreover, as soon as $h$ is upper-bounded on the
  support of $\mu$, the supremum with respect to $m$ on the right-hand
  side is reached for the Gibbs distribution $g$ given by
  \begin{equation*}
    \frac{\mathrm{d}g}{\mathrm{d}\mu}(a) =
    \frac{\exp\circ h(a)}{\int(\exp\circ\, h)\mathrm{d}\mu}, \quad a\in A.
  \end{equation*}
\end{lemma}

\FloatBarrier

\begin{proof}[Proof of \autoref{T1}]
Assume that \eqref{eq:condition} holds. For any $\e\in(0,1)$, we have
\begin{equation*}
\E\exp\left[ \delta(\ern(s)-\er(s))-\psi(s)-\log(1/\e)\right]\leq \e.
\end{equation*}
Now, assume that $s$ is drawn from some prior distribution $\pi$. We
can integrate on both sides of the previous inequality, thus
\begin{equation*}
\int \left\{\E\exp\left[ \delta(\ern(s)-\er(s))-\psi(s)-\log(1/\e)\right]\right\}\pi(\mathrm{d}s)\leq \e.
\end{equation*}
Using a Fubini-Tonelli theorem, we may write
\begin{equation*}
\E\int \exp\left[ \delta(\ern(s)-\er(s))-\psi(s)-\log(1/\e)\right]\pi(\mathrm{d}s)\leq \e.
\end{equation*}
Now, let $\rho$ denote an absolutely continuous distribution with
respect to $\pi$. We obtain
\begin{equation*}
\E\int\exp\left[ \delta(\ern(s)-\er(s))-\psi(s)-\log\frac{\mathrm{d}\rho}{\mathrm{d}\pi}(s)-\log(1/\e)\right]\rho(\mathrm{d}s)\leq \e.
\end{equation*}
With a slight extension of previous notation, we let $\E$ stands for the
expectation computed with respect to the distribution of $(\bX,Y)$
\emph{and} the posterior distribution $\rho$. Using the elementary inequality $\exp(\delta x)\geq
\1_{\R_+}(x)$, 
\begin{equation*}
  \PP\left[\ern(s)-\er(s)-\frac{\psi+\log(1/\e)+\log\frac{\mathrm{d}\rho}{\mathrm{d}\pi}(s)}{\delta} >0\right]\leq \e,
\end{equation*}
\ie,
\begin{equation}\label{pt-div}
  \PP\left[ \ern(s)\leq\er(s)+\frac{\psi(s)+\log(1/\e)+\log\frac{\mathrm{d}\rho}{\mathrm{d}\pi}(s)}{\delta} \right]\geq 1-\e.
\end{equation}
We may now apply the same scheme of proof with the variables
$\widetilde{T}_{i,j}=-T_{i,j}$ to obtain
\begin{equation*}
  \PP\left[ \er(s)\leq\ern(s)+\frac{\psi(s)+\log(1/\e)+\log\frac{\mathrm{d}\rho}{\mathrm{d}\pi}(s)}{\delta} \right]\geq 1-\e.
\end{equation*}
Now, for the choice $\rho=\hat{\rho}_\delta$ and $\hat{s}\sim\hat{\rho}_\delta$, 
\begin{equation*}
  \PP\left[ \er(\hat{s})\leq\ern(\hat{s})+\frac{\psi(\hat{s})+\log(1/\e)+\log\frac{\mathrm{d}\hat{\rho}_{\delta}}{\mathrm{d}\pi}(\hat{s})}{\delta} \right]\geq 1-\e.
\end{equation*}
Note that
\begin{align*}
  \log\frac{\mathrm{d}\hat{\rho}_{\delta}}{\mathrm{d}\pi}(\hat{s}) &=  \log\frac{\exp(-\delta L_n(\hat{s}))}{\int\exp(-\delta L_n(s^\prime))\pi(\mathrm{d}s^\prime)} =-\delta
  L_n(\hat{s})-\log\int\exp\left(-\delta L_n(s^\prime)\right)\pi(\mathrm{d}s^\prime) .
\end{align*}
Hence
\begin{equation*}
  \PP\left[\er(\hat{s})\leq-L_n(\eta)+\frac{1}{\delta}\left(
   \psi(\hat{s})+\log(1/\e)-\log\int\exp(-\delta L_n(s^\prime))\pi(\mathrm{d}s^\prime) \right)\right]\geq 1-\e.
\end{equation*}
From \autoref{lemma:catoni}, we obtain
\begin{equation*}
  \PP\left[\er(\hat{s})\leq \underset{\rho\in\mp}{\inf}\
    \left\{ \int \ern(s)\rho(\mathrm{d}s) + \frac{\psi(\hat{s})+\log(1/\e)+\KL(\rho,\pi)}{\delta} \right\}\right]\geq 1-\e,
\end{equation*}
where $s\sim\rho$. So by integrating \eqref{pt-div},
\begin{multline}\label{debTh3}
  \PP\left[\er(\hat{s})\leq \underset{\rho\in\mp}{\inf}\
    \left\{ \int \er(s)\rho(\mathrm{d}s) +\int \frac{\psi(s)}{\delta} \rho(\mathrm{d}s) + \frac{\psi(\hat{s})+2\log(2/\e)+2\KL(\rho,\pi)}{\delta} \right\}\right] \\ \geq 1-\e,
\end{multline}
which is the desired result.
\end{proof}

\begin{proof}[Proof of \autoref{lemmacond}]
For some candidate function $s$ and any $i,j=1,\dots,n$,
define $$T_{i,j}=\1_{\{(s(\bX_{i})-s(\bX_{j}))(Y_{i}-Y_{j})<0\}}-\1_{\{(\eta(\bX_{i})-\eta(\bX_{j}))(Y_{i}-Y_{j})<0\}}.$$
Using results on U-statistics \citep[Hoeffding decomposition of U-statistics,][]{Ser80}, we may write, for any
$\gamma>0$,
\begin{align}\label{eq:ustat}
\E\exp &\left[\gamma \sum_{i\neq j}(T_{i,j}-\E T_{i,j}) \right] = \E
\exp\left[\frac{\gamma
    n(n-1)}{n!}\sum_{\pi}\frac{1}{n/2}\sum_{i=1}^{n/2}(T_{\pi(i),\pi(i+n/2)}-\E
T_{\pi(i),\pi(i+n/2)})\right] \nonumber \\
&\leq \E\exp\left[2\gamma (n-1)\sum_{i=1}^{n/2}(T_{i,i+n/2}-\E
T_{i,i+n/2}) \right],
\end{align}
where we used the Jensen's inequality. Next, using an independence
argument and Hoeffding's inequality applied to the random variable $T_{i,i+n/2}-\E
T_{i,i+n/2}\in(-\E T_{i,i+n/2},1-\E T_{i,i+n/2})$,
\begin{align*}
  \E\exp\left[\gamma\sum_{i\neq j}(T_{i,j}-\E T_{i,j}) \right] &=
  \prod_{i=1}^{n/2}\E\exp\left[2\gamma (n-1)(T_{i,i+n/2}-\E
T_{i,i+n/2}) \right] \\
  &\leq \prod_{i=1}^{n/2}\exp\left(\frac{\gamma^2(n-1)^2}{2}\right)=\exp\left(\frac{n\gamma^2(n-1)^2}{4}\right) =\exp(\psi),
\end{align*}
with $\psi=\delta^2/4n$ and $\delta=\gamma n(n-1)$. Note that $\psi$ does not depend on $s$.
Finally, note that
\begin{equation*}
  \E\exp\left[\gamma \sum_{i\neq j}(T_{i,j}-\E T_{i,j}) \right] = \E\exp\left[ \delta\left(\ern(s)-\er(s)\right)\right].
\end{equation*}
\end{proof}

\begin{proof}[Proof of \autoref{coro1KL}]
From \autoref{T1} and since $\psi=\delta^{2}/4n$ does not depend on $s$, considering $\rho_{\m}$ as a probability measure whose support is $\R^{|\m|_{0}}$,
\begin{equation*}
  \PP\left[\er(\hat{s})\leq \underset{\m}{\inf}\ \underset{\rho_\m}{\inf}\
    \left\{ \int \er(s)  \rho(\mathrm{d}s) + \frac{2\psi+2\log(2/\e)+2\KL(\rho_{\m},\pi)}{\delta} \right\}\right]\geq 1-\e.
\end{equation*}
Next, note that
\begin{align}\label{eq:KL-dev}
  \KL(\rho_\m,\pi) &=
  \KL(\rho_\m,\pi_\m)+|\m|_0M\log(1/\beta)+\log\binom{d}{|\m|_0}+\log\frac{1-\beta^{d+1}}{1-\beta}
  \nonumber \\
  &\leq \KL(\rho_\m,\pi_\m)+|\m|_0M\log(1/\beta)+|\m|_0\log\frac{de}{|\m|_0}+\log\frac{1}{1-\beta},
\end{align}
where we used the elementary inequality $\log\binom{d}{k}\leq k\log\frac{de}{k}$. Note that if we consider as distributions $\rho_\m$ uniform distributions on
$\ell_2$-balls centered in any $\theta\in\mathcal{B}_\m$ such that $\|\theta\|=1$, of radius
$t\in(0,1
)$, we obtain that
  $\KL(\rho_\m,\pi_\m) =|\m|_0\log(1/t).$
For some $\theta_0$ such that $\|\theta_{0}\|=1$,
\begin{align*}
  R(s_{\theta})  &= \E[\1_{\{(s_{\theta}(\bX)-s_{\theta}(\bX'))(Y-Y') < 0\}}] \\
	&=\E[\1_{\{(s_{\theta_{0}}(\bX)-s_{\theta_{0}}(\bX'))(Y-Y') < 0\}}] \\ &\qquad+ \E[\1_{\{(s_{\theta}(\bX)-s_{\theta}(\bX'))(Y-Y') < 0\}}-\1_{\{(s_{\theta_{0}}(\bX)-s_{\theta_{0}}(\bX'))(Y-Y') < 0\}}]\\
	&\leq R(s_{\theta_{0}}) + \PP[\mathrm{sign}(s_{\theta}(\bX)-s_{\theta}(\bX'))(Y-Y') )\neq \mathrm{sign}(s_{\theta_{0}}(\bX)-s_{\theta_{0}}(\bX'))(Y-Y')]\\
	&= R(s_{\theta_{0}}) + \PP[\mathrm{sign}(s_{\theta}(\bX)-s_{\theta}(\bX'))\neq \mathrm{sign}(s_{\theta_{0}}(\bX)-s_{\theta_{0}}(\bX'))]\\
	&\leq R(s_{\theta_{0}}) + 2c\|\theta-\theta_{0}\|,
\end{align*}
where we used \autoref{cond_dens} in the last inequality. Let $\rho_{\m,\theta_0,t}$ denote the uniform distribution on the $\ell_2$-ball centered in $\theta_0$ and of radius $t\in(0,1)$. From what precedes,
\begin{equation}\label{eq_dens}
  \int \er(s_{\theta}) \rho_{\m,\theta_0,t}(\mathrm{d}s) = R(s_{\theta_{0}})+ 2ct.
\end{equation}
Using the notation 
\begin{equation}\label{constante}
K= 2\left(|\m|_0M\log(1/\beta)+|\m|_0\log\frac{de}{|\m|_0}+\log\frac{1}{1-\beta}\right),
\end{equation}
we obtain
\begin{multline*}
\PP\left[\er(\hat{s})\leq \underset{\m}{\inf}\ \underset{\theta\in\mathcal{B}_\m, \|\theta\|=1}{\inf}\
\underset{t\in(0,1)}{\inf}\
\left\{\vphantom{\frac{1}{2}}R(s_{\theta_{0}}) + 2ct \right.\right. \\ \left.\left. + \frac{2\psi+2\log(2/\e)+2|\m|_0\log(1/t) + 2K}{\delta} \right\}\right] \geq 1-\e.
\end{multline*}
It can easily be seen that the function $t\mapsto 2ct + \log(1/t)/\delta$ is
upper-bounded at the point $t=1/(2c\delta)$. Therefore,
\begin{multline*}\PP\left[\er(\hat{s})\leq \underset{\m}{\inf}\ \underset{\theta\in\mathcal{B}_\m, \|\theta\|=1}{\inf}\
    \left\{ R(s_{\theta_{0}} ) + \frac{1+2\psi+2\log(2/\e)+2|\m|_0\log(2c\delta) +2K}{\delta} \right\}\right] \geq 1-\e.
\end{multline*}
Now, recalling that $\psi=\delta^{2}/4n$, we choose $\delta=\sqrt{n}$.
The desired result is then straightforward from the proof of \autoref{T1}.
\end{proof}

The following \autoref{lemma-variance} allows us to obtain a tighter right-hand term in \eqref{eq:condition}, therefore leading to a refined oracle inequality with a faster rate of convergence. Let us introduce the notation $\phi(u)=e^{u}-u-1$.

\begin{lemma}\label{lemma-variance}
Let $s$ be a scoring function and $(\bX,Y)$ and
$(\bX^\prime,Y^\prime)$ two pairs of independent random variables. Let
$T(s)=\1_{\{(s(\bX)-s(\bX^\prime))(Y-Y^\prime)<0\}}-\1_{\{(\eta(\bX)-\eta(\bX^\prime))(Y-Y^\prime)<0\}}$, and $\V(Z)$ denote the variance of a random variable $Z$. Let
\autoref{cond-margin} hold for some $\alpha\in(0,1)$, then
\begin{equation*}
\V(T(s))\leq \C (L(s)-L^{*})^{\frac{\alpha}{1+\alpha}},
\end{equation*}
where $\C$ is a constant. Hence, for any distribution of the random variables $(\bX,Y)$ satisfying \autoref{cond-margin} for some parameter $\alpha\in(0,1)$, \eqref{eq:condition} holds with
$\psi(s)=\frac{n}{2}\V(T(s))\phi\left(\frac{2\delta}{n}\right)$. 
\end{lemma}

\begin{proof}[Proof of \autoref{lemma-variance}]
First, note that
$\E T=L(s)-L^\star\geq 0.$
Let
\begin{equation*}
r=r(\bX,\bX^\prime)=\mathrm{sign}(s(\bX)-s(\bX^\prime)),\quad r^\star=r^\star(\bX,\bX^\prime)=\mathrm{sign}(\eta(\bX)-\eta(\bX^\prime)),\quad
Z=(Y-Y^\prime)/2.
\end{equation*}
With this notation,
\begin{align*}
\E T^2 &=\E\big[
\1_{\{r\neq Z\}}+\1_{\{r^\star\neq Z\}}-2\1_{\{r\neq
  Z\}}\1_{\{r^\star\neq Z\}}\big] \\
&=
\E\big[
\1_{\{r=1\}}\1_{\{Z=-1\}}+\1_{\{r=-1\}}\1_{\{Z=1\}}+\1_{\{r^\star=1\}}\1_{\{Z=-1\}}+\1_{\{r^\star=-1\}}\1_{\{Z=1\}}
 \\  
&\qquad -2\1_{\{r=1\}}\1_{\{r^\star=
  1\}}\1_{\{Z=-1\}}-2\1_{\{r=-1\}}\1_{\{r^\star=-1\}}\1_{\{Z=1\}}
\big].
\end{align*}
Next,
\begin{align*}
\E T^2&=
\E\big[
\1_{\{r=1\}}(1-\eta(\bX))\eta(\bX^\prime)
+\1_{\{r=-1\}}\eta(\bX)(1-\eta(\bX^\prime))
+\1_{\{r^\star=1\}}(1-\eta(\bX))\eta(\bX^\prime)
 \\ 
&\qquad +\1_{\{r^\star=-1\}}\eta(\bX)(1-\eta(\bX^\prime))
-2\1_{\{r=1\}}\1_{\{r^\star= 1\}}(1-\eta(\bX))\eta(\bX^\prime)
 \\ 
&\qquad -2\1_{\{r=-1\}}\1_{\{r^\star=-1\}}\eta(\bX)(1-\eta(\bX^\prime))
\big].
\end{align*}
Thus,
\begin{align*}
\E T^2
&=\E\big[
(1-\eta(\bX))\eta(\bX^\prime)(\1_{\{r=1\}}+\1_{\{r^\star=1\}}-2\1_{\{r=1\}}\1_{\{r^\star= 1\}})
 \\ 
&\qquad
+\eta(\bX)(1-\eta(\bX^\prime))(\1_{\{r=-1\}}+\1_{\{r^\star=1\}}-2\1_{\{r=-1\}}\1_{\{r^\star= -1\}})
\big] \\
&=\E\big[
\1_{\{r\neq r^\star\}} ((1-\eta(\bX))\eta(\bX^\prime) +
  \eta(\bX)(1-\eta(\bX^\prime)))\big] \\
&=\E\big[
\1_{\{r\neq r^\star\}} (\eta(\bX) +
  \eta(\bX^\prime)-2\eta(\bX)\eta(\bX^\prime))\big] \\
&\leq\frac{1}{2}\E\big[\1_{\{r\neq r^\star\}}\big] \leq\frac{1}{2} C (L(s)-L^\star)^{\alpha/(1+\alpha)}.
\end{align*}
Finally,
\begin{equation*}
\V(T)=\E T^2-(\E T)^2\leq \C
(L(s)-L^\star)^{\alpha/(1+\alpha)},
\end{equation*}
where $\C$ is a constant, which proves the first statement.

Recalling \eqref{eq:ustat}, with the notation $\phi(u)=e^{u}-u-1$ and Benett's inequality,
we get
\begin{align*}
\E\exp\left[\gamma\sum_{i\neq j}\left(T_{i,j}(s)-\E T_{i,j}(s)\right) \right] 
&\leq\prod^{n/2}_{i=1} \exp\left(\V\left(T_{i,i+n/2}(s)\right)\phi\left(2(n-1)\gamma\right)\right)\\&= \exp\left(\frac{n}{2}\V(T(s))\phi(2(n-1)\gamma)\right),
\end{align*}
with $\gamma= \frac{\delta}{n(n-1)}$, which yields
\begin{align*}
\E\exp\left[\delta (\ern(s)-\er(s))\right]\leq \exp(\psi),
\end{align*}
with
  $\psi=\frac{n}{2}\V(T(s))\phi\left(\frac{2\delta}{n}\right),$
achieving the second statement of \autoref{lemma-variance}.
\end{proof}

\begin{proof}[Proof of \autoref{coro2}]
Now, combining what precedes and \eqref{debTh3}, 
\begin{multline*}
  \PP\left[\er(\hat{s})\leq \underset{\rho\in\mp}{\inf}\
    \left\{ \int \er(s)\rho(\mathrm{d}s) + \int\frac{n\V(T(s))\phi\left(2\delta/n\right)}{2\delta} \rho(\mathrm{d}s) \right.\right. \\ \left.\left. + \frac{n\V(T(\hat{s}))\phi\left(2\delta/n\right)/2+2\log(2/\e)+2\KL(\rho,\pi)}{\delta} \right\}\right]\geq 1-\e.
\end{multline*}
Using the elementary inequality $\phi(x)/x\leq x$ for any $x\in(0,1)$
yields
\begin{multline*}
  \PP\left[\er(\hat{s})\leq \underset{\rho\in\mp}{\inf}\
    \left\{ \int \er(s)\rho(\mathrm{d}s) +\int\frac{2\delta\V(T(s))}{n} \rho(\mathrm{d}s)+ \frac{2\delta\V(T(\hat{s}))}{n} \right.\right. \\ \left.\left. +\frac{2\log(2/\e)+2\KL(\rho,\pi)}{\delta} \right\}\right] \geq 1-\e.
\end{multline*}
Now, using \autoref{lemma-variance}, 
\begin{multline*}
  \PP\left[\er(\hat{s})\leq \underset{\rho\in\mp}{\inf}\
    \left\{ \int \er(s)\rho(\mathrm{d}s) +\int \frac{2\delta \C\er(s)^{\frac{\alpha}{1+\alpha}}}{n}\rho(\mathrm{d}s)+\frac{2\delta \C\er(\hat{s})^{\frac{\alpha}{1+\alpha}}}{n} \right.\right. \\ \left.\left. +\frac{2\log(2/\e)+2\KL(\rho,\pi)}{\delta} \right\}\right]  \geq 1-\e.
\end{multline*}
Thus, for any $x\geq 0$,
\begin{multline*}
  \PP\left[\left(1-\frac{2\delta\C x}{n}\right)\er(\hat{s})\leq \underset{\rho\in\mp}{\inf}\
    \left\{ \int \left(1+\frac{2\delta\C x}{n}\right)\er(s)\rho(\mathrm{d}s) \right.\right. \\ \left.\left. +
     \int \frac{2\delta\C}{n}\left(\er(s)^{\frac{\alpha}{1+\alpha}}-x\er(s)\right)\rho(\mathrm{d}s)+
      \frac{2\delta\C}{n}\left(\er(\hat{s})^{\frac{\alpha}{1+\alpha}}-x\er(\hat{s})\right)+\frac{2\log(2/\e)+2\KL(\rho,\pi)}{\delta} \right\}\right] \\   \geq 1-\e.
\end{multline*}
Clearly, the function $t\mapsto t^{\frac{\alpha}{1+\alpha}}-xt$ is
upper bounded by
$x^{-\alpha}\frac{1}{1+\alpha}\left(\frac{\alpha}{1+\alpha}\right)^\alpha$, so
\begin{multline}\label{eq:pourKL}
  \PP\left[\left(1-\frac{2\delta\C x}{n}\right)\er(\hat{s})\leq \underset{\rho\in\mp}{\inf}\
    \left\{ \int \left(1+\frac{2\delta\C x}{n}\right)\er(s)\rho(\mathrm{d}s) \right. \right.\\ \left. \left. +
    \int  \frac{2\delta\C\C_\alpha}{n} x^{-\alpha}\rho(\mathrm{d}s)+\frac{2\delta\C\C_\alpha}{n} x^{-\alpha}
			+\frac{2\log(2/\e)+2\KL(\rho,\pi)}{\delta} \right\}\right]  \geq 1-\e,
\end{multline}
where
$\C_\alpha=\frac{1}{1+\alpha}\left(\frac{\alpha}{1+\alpha}\right)^\alpha$. Next,
we choose $x=\frac{n}{4\delta\C}$. The function $\delta\mapsto
\C_\alpha\left(\frac{4\C}{n}\right)^{1+\alpha}\delta^{1+\alpha}+\frac{4}{\delta}$
is upper-bounded at the point $\delta=\Upsilon
n^{\frac{1+\alpha}{2+\alpha}}$ where
$\Upsilon=2((1+\alpha)\C_\alpha)^{-\frac{1}{2+\alpha}}\left(2\C\right)^{-\frac{1+\alpha}{2+\alpha}}$.
Thus we obtain
\begin{multline*}
  \PP\left[\er(\hat{s})\leq \underset{\rho\in\mp}{\inf}\   \left\{ 3\int \er(s) \rho(\mathrm{d}s) + n^{-\frac{1+\alpha}{2+\alpha}}\left[\C_\alpha\left(\C\Upsilon\right)^{1+\alpha}+\Upsilon^{-1}\left(\log(2/\e)+\KL(\rho,\pi)\right)\right]
       \right\}\right]  \geq 1-\e.
\end{multline*}
\end{proof}

\begin{proof}[Proof of \autoref{KL}]
Note that from \eqref{eq:pourKL}, choosing $x=\frac{n}{4\delta\C}$,
we get
\begin{multline*}
  \PP\left[\er(\hat{s})\leq \underset{\m}{\inf}\ \underset{\rho_\m}{\inf}\
    \left\{3\int \er(s) \rho(\mathrm{d}s) +
      2\C_\alpha\left(\frac{4\delta\C}{n}\right)^{1+\alpha}
      +\frac{4\log(2/\e)+4\KL(\rho_\m,\pi)}{\delta}
    \right\}\right]  \geq 1-\e.
\end{multline*}
Now, recall the result from \eqref{eq:KL-dev}.
\begin{multline*}
  \PP\left[\er(\hat{s})\leq \underset{\m}{\inf}\ \underset{\theta\in\mathcal{B}_\m, \|\theta\|=1}{\inf}\ \underset{t\in (0,1)}{\inf}\
    \left\{ 3\int \er(s_{\theta}) \rho_{\m,\theta,t}(\mathrm{d}s) + 2\C_\alpha\left(\frac{4\delta\C}{n}\right)^{1+\alpha}
      +\frac{4\log(2/\e)}{\delta} \right. \right.\\ \left. \left. +4
\frac{|\m|_0\log(1/t)+|\m|_0M\log(1/\beta)+|\m|_0\log\frac{de}{|\m|_0}+\log\frac{1}{1-\beta}
}{\delta}
    \right\}\right]\geq 1-\e.
\end{multline*}
With the notation stated in \eqref{constante} and the result \eqref{eq_dens},
\begin{multline*}
  \PP\left[\er(\hat{s})\leq \underset{\m}{\inf}\ \underset{\theta\in\mathcal{B}_\m, \|\theta\|=1}{\inf}\ \underset{t\in (0,1)}{\inf}\
    \left\{  3\er(s_{\theta}) +4ct + 2\C_\alpha\left(\frac{4\delta\C}{n}\right)^{1+\alpha}
      +\frac{4\log(2/\e)}{\delta} \right. \right.\\ \left. \left. +4
\frac{|\m|_0\log(1/t)+K}{\delta}
    \right\}\right]\geq 1-\e.
\end{multline*}
Since $t\mapsto 4ct + \log(1/t)/\delta$ is
upper-bounded at the point $t=1/(4c\delta)$,
\begin{multline}\label{eq:klsrbg}
  \PP\left[\er(\hat{s})\leq \underset{\m}{\inf}\ \underset{\theta\in\mathcal{B}_\m, \|\theta\|=1}{\inf}\
    \left\{  3\er(s_{\theta})  + 2\C_\alpha\left(\frac{4\delta\C}{n}\right)^{1+\alpha}
      +\frac{4\log(2/\e)}{\delta} \right. \right.\\ \left. \left. +4
\frac{|\m|_0\log(4c\delta)+K}{\delta}
    \right\}\right]\geq 1-\e.
\end{multline}
 Finally, noticing that the function $\delta\mapsto
\C_\alpha\left(\frac{4\delta\C}{n}\right)^{1+\alpha}+\frac{1}{\delta}$ is
upper-bounded at the point
$\delta=n^{\frac{1+\alpha}{2+\alpha}}\left(\frac{1}{4\C}\right)^{\frac{1+\alpha}{2+\alpha}}\left(\frac{1}{\C_\alpha(1+\alpha)}\right)^{\frac{1}{2+\alpha}}$,
we obtain
\begin{multline*}
  \PP\vast[\er(\hat{s})\leq \underset{\m}{\inf}\ \underset{\theta\in\mathcal{B}_\m, \|\theta\|=1}{\inf}\
    \Bigg\{ 3\er(s_\theta) \\
      +n^{-\frac{1+\alpha}{2+\alpha}} (4K+4\log(2/\e)+\log(4cn^{\frac{1+\alpha}{2+\alpha}})) \C_\alpha^{\frac{1}{2+\alpha}}\left(\frac{1}{4\C}\right)^{-\frac{1+\alpha}{2+\alpha}}
        \left[(1+\alpha)^{-\frac{1+\alpha}{2+\alpha}}
      + (1+\alpha)^{\frac{1}{2+\alpha}}\right]
    \Bigg\}\vast] \\ \geq 1-\e.
\end{multline*}
\end{proof}
\begin{proof}[Proof of \autoref{Sobolev}]
  We denote by $\eta^\proj$ the projection of $\eta$ onto
  $\mathcal{S}_{\Theta}$.  
  Let $\Gamma_{\eta^\proj}=\left\{(\bx,\bx^\prime)|
    \left(\eta^\proj(\bx)-\eta^\proj(\bx^\prime)\right)\left(\eta(\bx)-\eta(\bx^\prime)\right)<0\right\}$. For
  any $\bX,\bX^\prime\in \Gamma_{\eta^\proj}$, we have that
  $|\eta(\bX)-\eta(\bX^\prime)|\leq |\eta^\proj(\bX)-\eta(\bX)| +
  |\eta^\proj(\bX^\prime)-\eta(\bX^\prime)|$.
Since \citep[see][]{CLV08}
\begin{equation*}
L\left(\eta^\proj\right)-L^\star=\E \left[|\eta(\bX)-\eta(\bX^\prime)|\1\{(\bX,\bX^\prime)\in \Gamma_{\eta^\proj}\}\right],
\end{equation*}
we obtain $L(\eta^\proj)-L^\star\leq 2\E
[|\eta^\proj(\bX)-\eta(\bX)|]$.
Next, using the statement \eqref{eq:KL-dev} combined with
\autoref{T1}, result \eqref{eq_dens} and the choice $\psi=\delta^2/4n$ (which holds whatever
the distribution of $(\bX,Y)$ may be), we get that
\begin{multline*}
\PP\left[\er(\hat{s})\leq \underset{\m}{\inf}\ \underset{\theta\in\mathcal{B}_\m, \|\theta\|=1}{\inf}\ \underset{t\in (0,1)}{\inf}\
    \left\{ \er(s_{\theta}) +2ct + \frac{\delta}{2n}+\frac{2\log(2/\e)+|\m|_0\log(1/t)+2K}{\delta} \right\}\right] \\ \geq 1-\e,
\end{multline*}
where
$K=|\m|_0M\log(1/\beta)+|\m|_0\log\frac{de}{|\m|_0}+\log\frac{1}{1-\beta}$.
Since the function $t\mapsto 2ct + \log(1/t)/\delta$ is
upper-bounded at the point $t=1/(2c\delta)$, hence
\begin{multline*}
\PP\left[\er(\hat{s})\leq \underset{\m}{\inf}\ \underset{\theta\in\mathcal{B}_\m, \|\theta\|=1}{\inf}\
    \left\{ \er(s_{\theta}) + \frac{\delta}{2n}+\frac{1+2\log(2/\e)+|\m|_0\log(2c\delta)+2K}{\delta} \right\}\right] \\ \geq 1-\e,
\end{multline*}
Now,
\begin{align*}
\underset{\theta\in\mathcal{B}_\m, \|\theta\|=1}{\inf}\
\er(s_\theta) = \er(\eta^\proj) 
\leq 2 \E|\eta^\proj(\bX)-\eta(\bX)| 
&\leq 2 \sqrt{\E|\eta^\proj(\bX)-\eta(\bX)|^2} \\
&\leq 2 \sqrt{\int \sum_{j\in S^\star}\sum_{k\geq M+1}|\theta^{\star}_{jk}|^2\phi^2_k(x)\mu(\mathrm{d}x)} \\
&\leq 2\sqrt{B}\sqrt{\sum_{j\in S^\star}\sum_{k\geq M+1}|\theta^\star_{jk}|^2},
\end{align*}
since the $\phi_k$'s are such that $\int |\phi_k(x)|\mu(\mathrm{d}x)\leq B$ where $B>0$ is a numerical constant. Using
the definition of the Sobolev ellipsoid, we obtain
\begin{equation*}
\underset{\theta\in\mathcal{B}_\m, \|\theta\|=1}{\inf}\
\er(s_\theta)\leq 2\sqrt{B|S^\star|_0\kappa}(1+M)^{-\tau}.
\end{equation*}
Hence
\begin{multline*}
\PP\left[\er(\hat{s})\leq \underset{\m}{\inf}\ 
    \left\{ 2\sqrt{B\kappa|S^\star|_0}(1+M)^{-\tau} + \frac{\delta}{2n}+\frac{|m|_0M\log(1/\beta)}{\delta} \right.\right. \\ \left.\left. +\frac{1+|\m|_0\log(2c\delta)+2\log(2/\e)+2K^\prime}{\delta} \right\}\right]  \geq 1-\e,
\end{multline*}
where $K^\prime = |\m|_0\log\frac{de}{|\m|_0}+\log\frac{1}{1-\beta}$.
Next, observe that we get rid of the remaining infimum by substituting $|S^\star|$ to $|\m|_0$.

The function $t\mapsto
2\sqrt{B\kappa|S^\star|_0}(1+t)^{-\tau}+|S^\star|_0\log(1/\beta)t/\delta$ is
upper-bounded at the point $t=\left(\frac{\sqrt{|S^\star|_0}\log(1/\beta)}{
  2\sqrt{B\kappa}\tau\delta}\right)^{-\frac{1}{1+\tau}}-1$, which yields
\begin{equation*}
	\PP\left[\er(\hat{s})\leq \left\{
\zeta\delta^{\frac{-\tau}{1+\tau}}+ \frac{\delta}{2n} +\frac{1+2\log(2/\e)+2K'|+|S^\star|_0\log(2c\delta)}{\delta} \right\}\right]\geq 1-\e,
\end{equation*}
where $\zeta=	\left(\left(2\sqrt{B\kappa}\right)^{\frac{1}{1+\tau}}|S^\star|_0^{\frac{1+2\tau}{2+2\tau}}\log(1/\beta)^\frac{\tau}{1+\tau}\left(\tau^{-\frac{\tau}{1+\tau}}+\tau^{\frac{1}{1+\tau}}\right) \right)$, and
$K^\prime =|S^\star|_0\log\frac{de}{|S^\star|_0}+\log\frac{1}{1-\beta}$.
Since $t\mapsto
\zeta t^{-\frac{\tau}{1+\tau}}+\frac{t}{2n}$ is
upper-bounded at the point $t=n^{\frac{\tau+1}{2\tau+1}}\left(\frac{2\zeta\tau }{1+\tau}\right)^{\frac{\tau+1}{2\tau+1}}$,
\begin{multline*}
	\PP\left[\er(\hat{s})\leq \left\{
\zeta\left(\frac{2\zeta\tau n}{1+\tau}\right)^{\frac{\tau+1}{2\tau+1}\times\frac{-\tau}{1+\tau}}+ \frac{1}{2n}\left(\frac{2\zeta\tau n}{1+\tau}\right)^{\frac{\tau+1}{2\tau+1}} \right.\right. \\ \left.\left. +\left(2\log(2/\e)+2K'+|S^\star|_0\log(2c\delta)\right)\left(\frac{2\zeta\tau n}{1+\tau}\right)^{-\frac{\tau+1}{2\tau+1}} \right\}\right]  \geq 1-\e.
\end{multline*}
Finally,
\begin{multline*}
	\PP\left[\er(\hat{s})\leq \left\{
\xi n^{-\frac{\tau}{2\tau+1}} +(2\log(2/\e)+2K'+|S^\star|_0\log(C_{1}n^{\frac{\tau+1}{2\tau+1}}))\left(\frac{2\zeta\tau n}{1+\tau}\right)^{-\frac{\tau+1}{2\tau+1}} \right\}\right]  \geq 1-\e,
\end{multline*}
where $\xi=\zeta^{\frac{\tau+1}{2\tau+1}}2^{-\frac{\tau}{2\tau+1}}\left(\left(\frac{\tau }{1+\tau}\right)^{-\frac{\tau}{2\tau+1}}+\left(\frac{\tau }{1+\tau}\right)^{\frac{\tau+1}{2\tau+1}}\right)$ and $C_{1}=2c\left(\frac{2\zeta\tau }{1+\tau}\right)^{\frac{\tau+1}{2\tau+1}}$.
\end{proof}

\begin{proof}[Proof of \autoref{Sobolev-MA}]
Observe that \citep[as written in][Proposition 4]{CRICML}
\begin{align*}
\underset{\theta\in\mathcal{B}_\m, \|\theta\|=1}{\inf}\
\er(s_\theta) &= \er(\eta^\proj) \\
&\leq 2\| \eta^\proj-\eta\|_\infty \PP\left[ \left(\eta^\proj(\bX)-\eta^\proj(\bX^\prime)\right)\left(\eta(\bX)-\eta(\bX^\prime)\right)<0 \right] \\
&\leq 2C \|\eta^\proj-\eta\|_\infty \er(\eta^\proj)^{\frac{\alpha}{1+\alpha}} \quad \textrm{(by \autoref{cond-margin}).}
\end{align*}
Hence,
\begin{align*}
\er(\eta^\proj) \leq \left(2C \|\eta^\proj-\eta\|_\infty\right)^{1+\alpha} 
&\leq (2C)^{1+\alpha} \left(\underset{\bx}{\sup} \sum_{j\in S^\star}\sum_{k\geq M+1} \theta_{jk}\phi_k(x_j)\right)^{1+\alpha} \\
&\leq (2BC)^{1+\alpha} \left(\sum_{j\in S^\star}\sum_{k\geq M+1} |\theta_{jk}|\right)^{1+\alpha} \\
&\leq \left(2BC\sqrt{\kappa |S^\star|_0}\right)^{1+\alpha} (1+M)^{-\tau(1+\alpha)},
\end{align*}
by construction of the $\phi_k$'s.
Combining \eqref{eq:klsrbg} with what precedes, we get
\begin{multline*}
\PP\left[\er(\hat{s})\leq  2
    \left\{ \frac{3}{2} \left(2BC\sqrt{\kappa |S^\star|_0}\right)^{1+\alpha} (1+M)^{-\tau(1+\alpha)} + \left(\frac{4\C\delta}{n}\right)^{1+\alpha} \right. \right. \\ \left. \left.  +\frac{|S^\star|_0M\log(1/\beta)}{\delta}+\frac{2\log(2/\e)+2K'+|S^\star|_0\log(2c\delta)}{\delta} \right\}\right]\geq 1-\e,
\end{multline*}
The function $t\mapsto
3/2 \left(2BC\sqrt{\kappa |S^\star|_0}\right)^{1+\alpha} (1+t)^{-\tau(1+\alpha)}+|S^\star|_0\log(1/\beta)t/\delta$ is
upper-
bounded at the point $t=
\left(\frac{2|S^\star|_0 \log(1/\beta)}{
  3\tau(1+\alpha)\left(2BC\sqrt{\kappa |S^\star|_0}\right)^{1+\alpha}\delta}\right)^{-\frac{1}{1+\tau(1+\alpha)}}-1$, so we may write 
\begin{equation*}
	\PP\left[\er(\hat{s})\leq 2\left\{
\zeta\delta^{\frac{-\tau(1+\alpha)}{1+\tau(1+\alpha)}}+ \left(\frac{4\C\delta}{n}\right)^{1+\alpha} +\frac{2\log(2/\e)+2K'+|S^\star|_0\log(2c\delta)}{\delta} \right\}\right]\geq 1-\e,
\end{equation*}
where
\begin{multline*}
\zeta=	
\left(3/2 \left(2BC\sqrt{\kappa |S^\star|_0}\right)^{1+\alpha}\right)^{\frac{1}{1+\tau(1+\alpha)}} \left(\log(1/\beta)|S^\star|_0\right)^{\frac{\tau(1+\alpha)}{1+\tau(1+\alpha)}} \\ \times \left((\tau(1+\alpha))^{-\frac{\tau(1+\alpha)}{1+\tau(1+\alpha)}}+(\tau(1+\alpha))^{\frac{1}{1+\tau(1+\alpha)}}\right).
\end{multline*}
The function $t\mapsto
\zeta t^{-\frac{\tau}{1+\tau(1+\alpha)}}+\left(\frac{4\C t}{n}\right)^{1+\alpha}$ (it is sufficient to consider this part) is
upper-bounded at the point
$t=\left(\frac{\zeta \tau(1+\alpha)}{1+\tau(1+\alpha)}\right)^{\frac{1+\tau(1+\alpha)}{(1+\tau(2+\alpha))(1+\alpha)}} \left(\frac{n}{4\C}\right)^{\frac{1+\tau(1+\alpha)}{1+\tau(2+\alpha)}},
$
hence using the notation $\Upsilon=\left(\frac{\zeta \tau(1+\alpha)}{1+\tau(1+\alpha)}\right)^{\frac{1+\tau(1+\alpha)}{(1+\tau(2+\alpha))(1+\alpha)}} \left(\frac{1}{4\C}\right)^{\frac{1+\tau(1+\alpha)}{1+\tau(2+\alpha)}}$,
we obtain that
\begin{multline*}
	\PP\left[\er(\hat{s})\leq 2\left\{
\left(\zeta \Upsilon^{-\frac{\tau(1+\alpha)}{1+\tau(1+\alpha)}}+(4\C \Upsilon)^{1+\alpha}\right)n^{\frac{-\tau(1+\alpha)}{1+(2+\alpha)\tau}} \right.\right. \\ \left.\left. +\Upsilon^{-1}n^{\frac{-(1+\tau(1+\alpha))}{1+(2+\alpha)\tau}}(2\log(2/\e)+2K'+|S^\star|_0\log(2c\Upsilon n^{\frac{1+\tau(1+\alpha)}{1+\tau(2+\alpha)}}))
 \right\}\right]  \geq 1-\e.
\end{multline*}
\end{proof}

\section*{Acknowledgements}

The authors would like to thank the Associate Editor and a Reviewer for valuable comments and insightful suggestions, which led to a substantial improvement of the paper.

\bibliographystyle{abbrvnat}
\footnotesize{
\bibliography{PACRanking}
}
\end{document}